\documentclass[journal]{IEEEtran}
\usepackage{lineno,hyperref}
\hypersetup{hidelinks,
	colorlinks=true,
	allcolors=black,
	pdfstartview=Fit,
	breaklinks=true}
\usepackage{bbm}
\usepackage{bm}
\usepackage{amsfonts}
\usepackage{amsmath}
\usepackage{mathrsfs}
\usepackage{amssymb}
\usepackage{enumerate}
\usepackage{graphicx}
\usepackage{subfigure}
\usepackage{booktabs}
\usepackage{graphicx}
\usepackage{setspace}
\usepackage{algorithm}
\usepackage{algorithmic}
\usepackage{setspace}
\usepackage{multirow}
\usepackage{color}
\usepackage{cite}
\usepackage{array}
\usepackage{times}
\usepackage{mathptmx}
\usepackage{dsfont}
\ifCLASSINFOpdf
\else
\fi
\begin{document}
\title{CHITNet: A Complementary to Harmonious Information Transfer Network for Infrared and Visible Image Fusion}
\author{Keying Du, Huafeng Li, Yafei Zhang, Zhengtao Yu  
\thanks{This work was supported in part by the National Natural Science Foundation of China under Grant 62161015, and the Yunnan Fundamental Research Projects (202301AV070004).}
\thanks{K. Du, H. Li, Y. Zhang and Z. Yu are with the Faculty of Information Engineering and Automation, Kunming University of Science and Technology, Kunming 650500, China.(E-mail:zyfeimail@163.com (Y. Zhang))}
\thanks{}
\thanks{Manuscript received xxxx;}}
\markboth{Journal of \LaTeX\ Class Files}%
{Shell \MakeLowercase{\textit{et al.}}}
\maketitle
\begin{abstract}
Current infrared and visible image fusion (IVIF) methods go to great lengths to excavate complementary features and design complex fusion strategies, which is extremely challenging. To this end, we rethink the IVIF outside the box, proposing a complementary to harmonious information transfer network (CHITNet). It reasonably transfers complementary information into harmonious one, which integrates both the shared and complementary features from two modalities. Specifically, to skillfully sidestep aggregating complementary information in IVIF, we design a mutual information transfer (MIT) module to mutually represent features from two modalities, roughly transferring complementary information into harmonious one. Then, a harmonious information acquisition supervised by source image (HIASSI) module is devised to further ensure the complementary to harmonious information transfer after MIT. Meanwhile, we also propose a structure information preservation (SIP) module to guarantee that the edge structure information of the source images can be transferred to the fusion results. Moreover, a mutual promotion training paradigm with interaction loss is adopted to facilitate better collaboration among MIT, HIASSI and SIP. In this way, the proposed method is able to generate fused images with higher qualities. Extensive experimental results demonstrate the superiority of CHITNet over state-of-the-art algorithms in terms of visual quality and quantitative evaluations. The source code of the proposed method can be available at {\url{https://github.com/lhf12278/CHITNet}}.
 
\end{abstract}
\IEEEpeerreviewmaketitle
\section{Introduction}
Visible light imaging sensors can clearly image objects visible to the human eye, whose imaging results have the advantages of high spatial resolution, rich texture details, and more in line with human visual perception. However, such sensors fail to clearly image objects in severe weather scenes such as fog and haze. Infrared sensors image through measuring the temperature difference between target and background, able to effectively avoid the influence of harsh conditions on target imaging, yet still cannot clearly image the background information with a relatively constant temperature \cite{1}. In order to integrate the advantages of those two types of imaging sensors, infrared and visible image fusion (IVIF) technology is proposed, which can combine the mutual information extracted by the two types of sensors into one image with both clear target and background, achieving a more comprehensive description of the scene. Currently, this technology has been widely used in space exploration, resource exploration, biomedicine, equipment detection, target tracking and other fields.

In recent years, a series of research achievements have been made in infrared and visible image fusion. Effective fusion methods have emerged in large numbers, which can be roughly divided into multi-scale transform-based \cite{2,49,54,55}, sparse representation-based \cite{45,46,47,50}, and deep learning-based fusion methods \cite{48,52,75,53,56,59,60}. Fusion methods based on multi-scale transform generally perform multi-scale transformation on the source images, then fuse the transform coefficients, and use multi-scale inverse transformation to reconstruct the fusion result. However, multi-scale transformation involved in this type of method often represents the image with a fixed basis, which has weak sparsity and limits the further improvement of fusion performance. In contrast, the sparse representation-based fusion methods can effectively alleviate the above problems by building an over complete dictionary for the represented images from a set of training samples. Nevertheless, it still has weak performance in mining the statistical characteristics of large samples. Deep learning-based methods enjoy high favor among researchers lately because they can effectively solve this problem.

\begin{figure}[t!]
	\centering
	\includegraphics[width=0.5\textwidth]{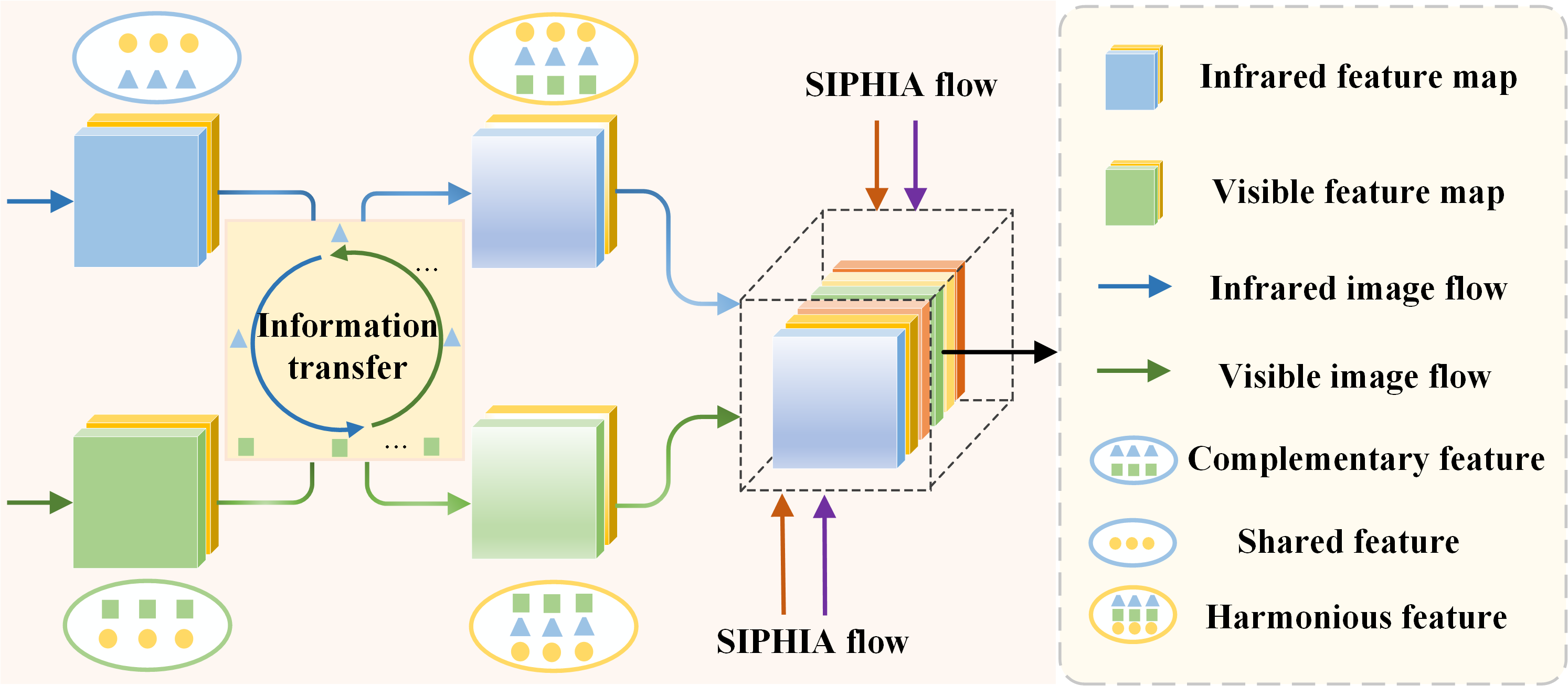}
	\caption{The core idea of our proposed CHITNet. Harmonious feature indicates the more comprehensive ones that cover shared information of infrared and visible images and modality-specific information from both two modal source images.}
	\label{label1}
\end{figure}

At present, numerous excellent deep learning-based IVIF methods have emerged, which are all committed to transferring complementary information from infrared and visible images to the fusion results. Achieving this goal generally requires to make two efforts. On the one hand, the feature extraction network is expected to have excellent performance in complementary information extraction. However, it is extremely challenging for the feature extraction network to effectively perceive complementary information and pay more attention to it. On the other hand, complex fusion strategies are often designed to aggregate the complementary information. In addition, existing fusion schemes all regard harmonious information of the source images as meaningless, and strive to ignore it. Different from the above methods, in order to alleviate challenge posed by the complementary information excavation of feature extraction network, this paper rethinks the deep learning-based fusion idea, that is, complementary to harmonious information transfer, which can effectively avoid weak feature loss of the source images, and transfer those weak details to the fusion results. Specifically, the core idea of the proposed method is shown in Fig. \ref{label1}. We consider that the infrared and visible images of the same scene cover both shared and modality-specific (complementary) features. If the infrared-specific features are transferred into the visible one, we can obtain features containing shared, infrared-specific and visible-specific information, $i.e.$ harmonious features, and vice versa. In other words, harmonious features are those more comprehensive ones that cover both modality-shared and modality-specific information. Compared to the complementary feature based fusion, harmonious feature based fusion can not only reduce the burden of feature extraction network mining complementary information, but also avoid the complementary information loss in the fusion process, getting rid of the reliance on complex fusion schemes.

Based on the above ideas, we propose a dedicated complementary to harmonious information transfer network (CHITNet) consisting of mutual information transfer (MIT) module, structure information preserved harmonious information acquisition (SIPHIA) module and fusion result reconstruction (FRR) module. MIT is composed of three parts: image inversion (II), feature extraction (FE) and mutual representation learning (MRL). II mainly facilitates the feature extraction network to obtain richer information. FE is responsible for extracting features from the input images for subsequent information transfer and fusion. MRL realizes interactive information transmission between infrared and visible source images, achieving complementary to harmonious information transfer. SIPHIA contains harmonious information acquisition supervised by source image (HIASSI) module and structure information preservation (SIP) module. HIASSI ensures that the complementary information after MRL can become harmonious one. SIP promises that the edge structure information of the source images can be transferred to the fusion result. FRR constructs the final fusion image. The above design can effectively alleviate complementary feature extraction challenge, and meanwhile avoid designing complex fusion mechanism to aggregate complementary information. Furthermore, similar to \cite{44}, we adopt a mutual promotion training paradigm (MPTP) between two SIPHIAs and the MIT. Concretely, we divide the whole training process to two circulative phases, in other words they are trained in turn. In the first cycle phase, we train MIT to have the basic ability of information transfer by performing interactive feature representation. And then in the second cycle phase, we train two SIPHIAs to create impressive edge information for fusion, treating the fusion results obtained by MIT as “labels” to further improve the harmonious feature and edge details extraction ability of two SIPHIAs. In summary, the main contributions of the proposed method are as follows:

\begin{itemize}
	\item We reconsider the IVIF task from a new aspect, and propose an innovative idea, which transfers complementary information into harmonious one to avoid the tricky complementary information extraction and aggregation problem.
	
	\item Benefited from this interesting notion, we exploit an efficient IVIF model CHITNet to realize complementary to harmonious information transfer in source infrared and visible images. Meanwhile, a feature protection mechanism is implanted, migrating detailed edge information to the fusion result.
	
	\item Extensive experiments conducted on three publicly available datasets demonstrate the effectiveness of our method, as well as show its superiority over other state-of-the-art (SOTA) models.
\end{itemize}

The rest of this article is organized as follows: Section \uppercase\expandafter{\romannumeral2} discusses typical deep learning based and high-level task-driven IVIF methods. In Section \uppercase\expandafter{\romannumeral3}, we describe the details of the proposed method. Section \uppercase\expandafter{\romannumeral4} shows the experimental results and analysis, and Section \uppercase\expandafter{\romannumeral5} summarizes the work and draws some conclusions.

\section{Related Work}
The IVIF task has been extensively explored for years and attracted much attention from the community due to its wide application prospects. In this section, we will briefly review the existing classical and deep learning based IVIF algorithms that are closely related to ours and some high-level task-driven IVIF algorithms.

\subsection{Classical infrared-visible image fusion}
Classical fusion frameworks usually realize image fusion in the transform domain and spatial domain through designing appropriate feature extraction details and fusion rules, which generally contain two major categories, multi-scale transform-based methods \cite{49,64} and sparse representation-based methods \cite{66,46,50,48,47}.
Multi-scale transform-based methods first decompose source images into several levels, as the feature extraction, then fuse corresponding layers with particular rule, and reconstruct the target images accordingly, where popular transforms used for decomposition and reconstruction include wavelet \cite{67}, pyramid \cite{68}, curvelet \cite{69}, and their revised versions. However, these methods typically tend to leave out image details in the fused results and lead to halos or undesirable artifacts in the fused result due to the fixed bases used in the multi-scale transform-based methods. The key of sparse representation-based methods is to build over complete dictionaries from a large number of natural images to possibly represent the source images with linear combinations of sparse bases. Although sparse representation-based methods have achieved promising performance, a limited number of dictionaries cannot reflect the full information of input images, obscuring details such as edges and textures in the source images.

\subsection{Deep learning based infrared-visible image fusion}
\textbf{CNN-based methods}. In recent years, convolution neural network (CNN) dominates the field of image fusion in virtue of its strong adaptability. Commonly, the entire image fusion procedure includes feature extraction, feature fusion and image reconstruction. Some methods merely use CNN to realize activity level measurement and generate a weight map for hand-crafted features \cite{6,58}, while the main fusion process remains traditional ways. Other methods adopt end-to-end mode, carrying out every step in image fusion with CNN. For instance,  Zhang et al. \cite{7} explored a fast unified image fusion network based on proportional maintenance of gradient and intensity. They defined a unified loss function which can adapt to different fusion tasks according to hyperparameters. In order to better preserve the thermal targets in infrared images and the texture structures in visible images, Ma et al. \cite{8} introduced a salient target mask to annotate desired regions. Their network can achieve salient target detection and fuse key information in an implicit manner. Moreover, Xu et al. \cite{9} integrated different fusion tasks into a unified framework with adaptive information preservation degrees. Xu et al. \cite{74} presented a new unsupervised and unified densely connected network for different types of image fusion tasks. Zhang et al. \cite{72} proposed to realize multi-modal and digital photography image fusion in real time. Li et al. \cite{76} proposed a novel registration free
fusion method for the infrared and visible images with translational displacement. Recently, Zhao et al. \cite{73} proposed a new paradigm for end-to-end self-supervised learning to tackle the challenge of effective fusion model training due to the scarcity of ground truth fusion data. Obviously, the design of these complex fusion strategies is a big problem.

\textbf{AE-based methods}. Given that the auto-encoder (AE) structure can learn effective feature representation in an unsupervised manner, AE-based methods are utterly suitable for IVIF task since there commonly lacks fused ground-truth. Most of them trained an auto-encoder to achieve both feature extraction and image reconstruction. Typically, Li et al. \cite{10} presented a novel deep learning architecture with encoding network combined by convolutional layers, fusion layer and dense block, while the fused image was reconstructed by decoder. Furthermore, they introduced NestFuse \cite{11} and RFN-Nest \cite{12} to preserve significant amounts of information from input data in a multi-scale perspective. Specifically, the latter designed the detail preservation and feature enhancement loss functions to integrate more useful information from source images. To address the limitation of the uninterpretability of deep feature maps, Xu et al. \cite{13} proposed an unsupervised deep learning method to realize the interpretable importance evaluation of feature maps. Liang et al. \cite{70} proposed a powerful image decomposition model for fusion task via the self-supervised representation learning, which can decompose the source images into a feature embedding space, where the common and unique features can be separated without any paired data or sophisticated loss. Nevertheless, the auto-encoder is trained by minimizing the reconstruction loss, which may lead the model to focus too much on copying the input data rather than extracting the essential features. When the training dataset contains noise or redundant information, the auto-encoder is inclined to learn these noise features, largely affecting the generalization ability of the model.

\textbf{GAN-based methods}. Thanks to the strong ability to estimate probability distributions, the generative adversarial network (GAN) is ideal for unsupervised tasks such as image fusion. Ma et al. \cite{14} played an adversarial game between a generator and a discriminator for the first time, keeping the infrared thermal radiation information and preserving the visible appearance texture information. Ma et al. \cite{71} proposed an end-to-end model for infrared and visible image fusion based on detail preserving adversarial learning, which is able to overcome the limitations of the manual and complicated design of activity-level measurement and fusion rules in traditional fusion methods. Unfortunately, a single discriminator is prone to lead the fused image to be similar to only one of the source images, causing the loss of some information. Therefore, DDcGAN \cite{15}, a dual-discriminator conditional generative adversarial network was designed to simultaneously keep the most important feature information in infrared and visible images. Besides, the existing GAN-based infrared and visible image fusion methods fail to highlight the typical regions in infrared and visible images because of the inability to perceive the most discriminative parts. To this end, AttentionFGAN \cite{16} integrated multi-scale attention mechanism into both generator and discriminator of GAN to capture more typical regions. Subsequently, GANMcC \cite{17} was devised to transform image fusion into muti-distribution simultaneous estimation problem, which made the fused results have both the distributions of infrared and visible domains in a more balanced manner. However, GAN-based methods are naturally faced with problems such as training instability, model crash, data overfitting, high computational cost, and difficulty in judging convergence, etc.

In addition, above mentioned deep learning-based approaches put emphasis on extracting and integrating the complementary information from the infrared and visible images, which is an extremely hard problem. To this end, we rethink the IVIF task from a brand new aspect, focusing more on the harmonious information excavation which is relatively easier to achieve. Besides, we do not struggle to design complex fusion strategies, but use simple while direct operation to achieve effective fusion.

\subsection{High-level task-driven infrared-visible image fusion}
Low-level vision tasks are often pre-processed and serve for downstream high-level vision tasks, $e.g.$ object detection, semantic segmentation, $etc.$. For the infrared and visible image fusion task, Tang et al. \cite{22} introduced the semantic loss to integrate more semantic information into fused images. They further incorporated image registration, image fusion, and semantic requirements of high-level vision tasks into a single framework and proposed a novel image registration and fusion method, named SuperFusion \cite{51}. Consequently, since object detection focuses on sparse object regions in images and is important in disaster rescue and traffic field, Sun et al. \cite{23} used detection-driven losses to provide task-related information for the detection-driven IVIF network. More recently, Liu et al. \cite{41} proposed a bilevel optimization formulation for the joint problem of fusion and detection, named TarDAL network for fusion and a commonly used detection network.

However, high-level vision task-driven image fusion often requires elaborately designed down-stream network to tackle the high-level vision task, which is relatively cumbersome and lacking in generality. Conversely, our proposed method is more universal without specific down-stream task oriented framework design.

\section{The Proposed Method}
\subsection{Overview}
Most existing IVIF methods are dedicated to mining the complementary information in two source images, which is quite challenging. Therefore, we rethink the problem from another new perspective, where we focus on mining the harmonious information in infrared and visible images, since such information is relatively easy to acquire and utilize in sequential fusion process. To be more specific, we propose a specialized complementary to harmonious information transfer framework to tactfully evade the complementary information extraction problem during infrared and visible image fusion.

\begin{figure*}[t!]
	\centering
	\includegraphics[width=\textwidth]{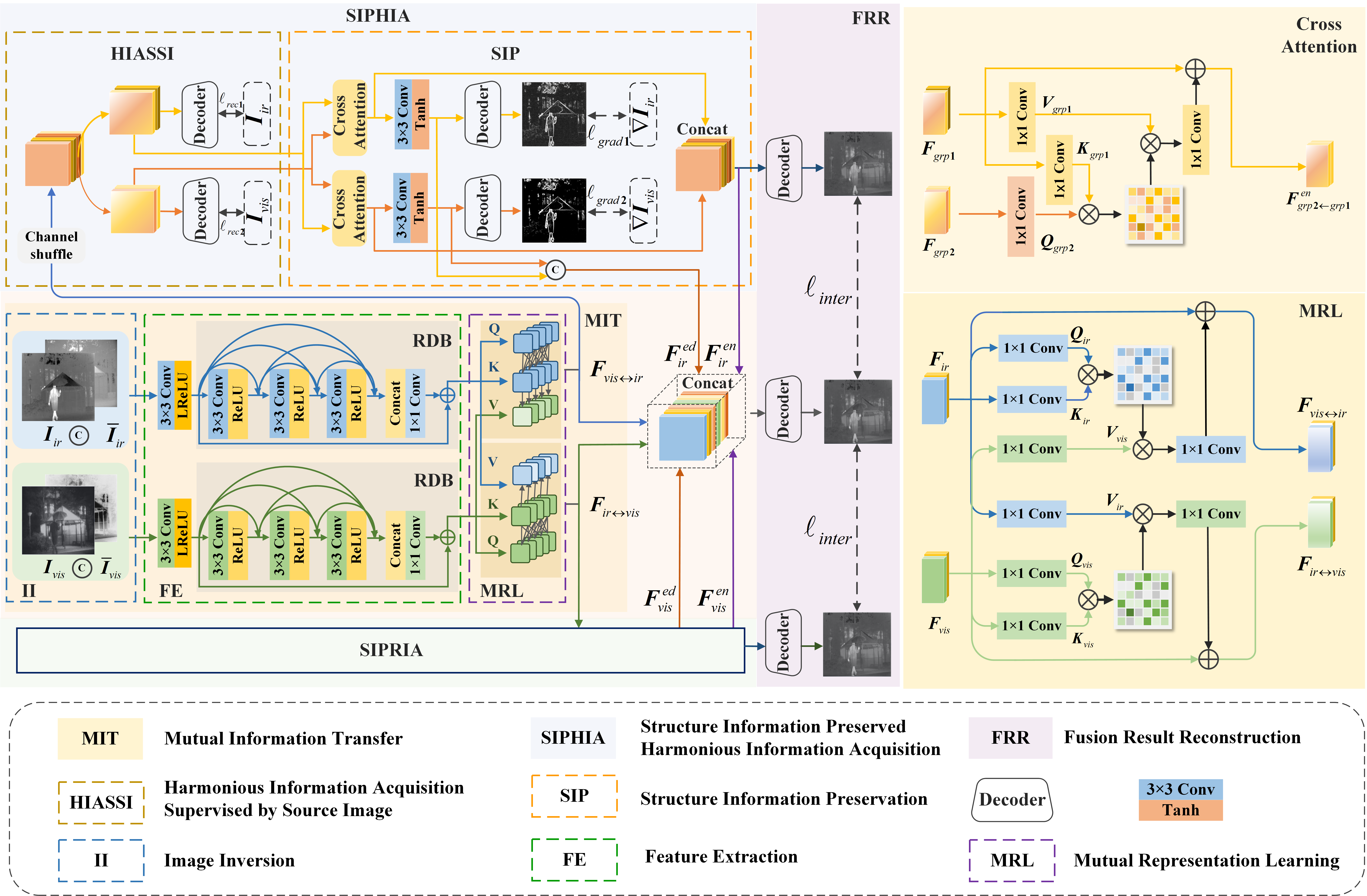}
	\caption{Overall architecture of the proposed method. In phaseM, the input concatenated infrared and visible image pairs $\bm I_{ir}^{\prime}$ and $\bm I_{vis}^{\prime}$ are fed into the infrared-visible feature encoder $\bm E_{ir}$ and $\bm E_{vis}$ respectively to obtain features $\bm F_{ir}$ and $\bm F_{vis}$. MIT is performed on $\bm F_{ir}$ and $\bm F_{vis}$ to achieve basic complementary to harmonious information transfer, getting transferred features $\bm F_{vis \leftrightarrow ir}$ and $\bm F_{ir \leftrightarrow vis}$. In phaseS, we then send $\bm F_{vis \leftrightarrow ir}$ and $\bm F_{ir \leftrightarrow vis}$ to SIPHIA and attain $\bm F_{ir}^{ed}$, $\bm F_{ir}^{en}$, $\bm F_{vis}^{ed}$, and $\bm F_{vis}^{en}$, ensuring successful information transfer and effectively preserving structure information. Finally, $\bm F_{vis \leftrightarrow ir}$, $\bm F_{ir}^{ed}$, $\bm F_{ir}^{en}$, $\bm F_{ir \leftrightarrow vis}$, $\bm F_{vis}^{ed}$, and $\bm F_{vis}^{en}$ are concatenated together and sent to the decoder $\bm D_{fuse}$ to obtain the final fusion result $\bm I_{fused}$.}
	\label{label2}
\end{figure*}

Pipeline of the proposed CHITNet is shown in Fig.~\ref{label2}, which consists of the mutual information tranfer (MIT) module, the structure information preserved harmonious information acquisition (SIPHIA) module and the fusion result reconstruction (FRR) module. MIT mutually represents features from two modalities, roughly transferring complementary information into harmonious one. SIPHIA aims to extract harmonious as well as abundant information, especially edge structure details from source images of two modalities. Besides, it is noteworthy that we do not adopt the end-to-end training paradigm but choose muti-stage mutual promotion training strategy to achieve effective interaction and co-reinforcement between MIT and SIPHIA.

\subsection{Mutual information transfer}\label{section:3.2}
We design the mutual information transfer (MIT) module to realize complementary to harmonious information transfer. MIT consists of three key components, namely image inversion (II), feature extraction (FE), and mutual representation learning (MRL). Below we describe the concrete structures of II, FE and MRL.
\subsubsection{Image inversion and feature extraction}
To make better use of the content in relatively over-exposed areas, literature \cite{24} inverted the source image so that the originally over-exposed regions would appear like underexposed ones in the inverted image. In light of the infrared and visible image characteristics, we find that the salient targets in grayscale inverted image tend to resemble that in the corresponding image from another modality, $i.e.$, the salient target in inverted infrared image is alike with that in its corresponding visible image, and so does the inverted visible image. Thus, concatenating the original image and its inverted image will facilitate the complementary to harmonious information transfer. Given source images $\bm I_{ir}$ and $\bm I_{vis}$, we invert them by $\overline{\bm I}_{ir}=1- \bm I_{ir}$ and $\overline{\bm I}_{vis}=1-\bm I_{vis}$. Pairs of $\bm I_{ir}$ and $\overline{\bm I}_{ir}$ are concatenated in channel dimensions to obtain $\bm I_{ir}^{\prime}$, and pairs of $\bm I_{vis}$ and $\overline{\bm I}_{vis}$ are concatenated to obtain $\bm I_{vis}^{\prime}$. $\bm I_{ir}^{\prime}$ and $\bm I_{vis}^{\prime}$ are fed into the infrared-visible feature encoder $\bm E_{ir}$ and $\bm E_{vis}$, respectively. The extracted infrared and visible image features are presented by 
\begin{equation}
	\begin{aligned}
		\bm F_{ir} &= \bm E_{ir}(\bm I_{ir}^{\prime}), \\
		\bm F_{vis} &= \bm E_{vis}(\bm I_{vis}^{\prime}),
	\end{aligned}
\end{equation}
where $\bm E_{ir}$ and $\bm E_{vis}$ have the same structure consisting of shallow feature extraction and a residual dense block but do not share parameters, which follows the residual dense network \cite{43}.

Detailed structure of the feature extraction process is shown in Fig.~\ref{label2}, whose mathematical expressions can be formulated as:
	\begin{equation}
		\begin{aligned}
			\bm	F_{ir}^s &= {\rm{LReLU}}\left( {Con{v_{3 \times 3}}\left( {{\bm  {I'}_{ir}}} \right)} \right),\\
			\bm	F_{vis}^s &= {\rm{LReLU}}\left( {Con{v_{3 \times 3}}\left( {{\bm {I'}_{vis}}} \right)} \right),
		\end{aligned}
	\end{equation}
	where LReLU is leaky rectified linear units (LeakyReLU), and $Con{v_{3 \times 3}}$  denotes $3 \times 3$ convolution operation to extract shallow features.
	\begin{equation}
		\begin{aligned}
			\bm F_{ir} &= RDB\left( {F_{ir}^s} \right) \\
			\bm	F_{vis} &= RDB\left( {F_{vis}^s} \right),
		\end{aligned}
	\end{equation}
	where $RDB\left(  \cdot  \right)$ denotes the operations of the residual dense block, which can be a composite function of operations, such as convolution and rectified linear units (ReLU).

\subsubsection{Mutual representation learning}
The architecture of MRL module is presented in Fig. \ref{label2}, which strives to transfer complementary information hidden in source images to harmonious one through mutual modality feature representation. 

Similar to attention mechanism in Transformer \cite{25}, the infrared image feature $\bm F_{ir} \in {\mathbb{R}^{C \times H \times W}}$ (where $H$ and $W$ are the height and width, $C$ is the number of channels) and the visible image feature $\bm F_{vis} \in {\mathbb{R}^{C \times H \times W}}$ pass through feature vectorization operations and convolutions to generate Query ${{\bf{Q}}_{ir}} \in {\mathbb{R}^{C \times HW}}$, Key ${{\bf{K}}_{ir}} \in {\mathbb{R}^{C \times HW}}$, Value ${{\bf{V}}_{ir}} \in {\mathbb{R}^{C \times HW}}$, and Query ${{\bf{Q}}_{vis}} \in {\mathbb{R}^{C \times HW}}$, Key ${{\bf{K}}_{vis}} \in {\mathbb{R}^{C \times HW}}$, Value ${{\bf{V}}_{vis}} \in {\mathbb{R}^{C \times HW}}$, respectively.
Given $\bm F_{ir}$ and $\bm F_{vis}$, their mutual feature representations are obtained through information transfer:
\begin{equation}
	\begin{aligned}
		{\widetilde{\bm F}_{vis \leftrightarrow ir}} &=  {\mathop{\rm softmax}\nolimits} (\frac{{{{\bf{Q}}_{ir}}{{({{\bf{K}}_{ir}})}^T}}}{{\sqrt d }}){{\bf{V}}_{vis}},\\
		{\widetilde{\bm F}_{ir \leftrightarrow vis}} &= {\mathop{\rm softmax}\nolimits} (\frac{{{{\bf{Q}}_{vis}}{{({{\bf{K}}_{vis}})}^T}}}{{\sqrt d }}){{\bf{V}}_{ir}},
	\end{aligned}
\end{equation}
where $\leftrightarrow$ is cross-modality mutual representation, ${\sqrt d }$ is a normalization factor, and $T$ is transpose operation. ${{{\bf{Q}}_{ir}}{{({{\bf{K}}_{ir}})}^T}}$ refers to global information representation of $\bm F_{ir}$, multiplying with ${\bf{V}}_{vis}$ to obtain visible modality representation of $\bm F_{ir}$, and vice versa. Thus we can achieve cross-modality mutual representation. To further enhance the gained interactive feature representation and avoid information loss, we perform the following operations:
\begin{equation}
	\begin{aligned}
		\bm F_{vis \leftrightarrow ir} &= Conv_{1\times1}(\widetilde{\bm F}_{vis \leftrightarrow ir}) \oplus {\bm F_{ir}},\\
		\bm F_{ir \leftrightarrow vis} &= Conv_{1\times1}(\widetilde{\bm F}_{ir \leftrightarrow vis}) \oplus {\bm F_{vis}},
	\end{aligned}
\end{equation}
where $Conv_{1\times1}$ is $1 \times 1 $ convolution. $\oplus$ denotes summing operation.
\subsection{Structure information preserved harmonious information acquisition}\label{section:3.3}
Infrared SIPHIA generates $\bm F_{ir}^{en}$ and $\bm F_{ir}^{ed}$. Similarly, visible SIPHIA generates $\bm F_{vis}^{en}$ and $\bm F_{vis}^{ed}$. $\bm F_{m}^{en}$ $(m \in {ir,vis})$ are enhanced features added to the final fusion process, avoiding information loss, while $\bm F_{m}^{ed}$ are edge detail features injected to the fused results to help with the structure information preservation. SIPHIA mainly consists of HIASSI and SIP, which will be demonstrated in detail as follows. Infrared SIPHIA and visible SIPHIA are weight-specific.
\subsubsection{Harmonious information acquisition supervised by source image}
To further ensure the complementary to harmonious information transfer after mutual representation learning, HIASSI is proposed to enable the features injected to the fusion process to have abundant information from both infrared and visible modalities by elaborately designing image reconstruction, so as to maximize the redundancy of the whole network. The specific structure of HIASSI is shown in Fig. \ref{label2}. $\bm F_{vis \leftrightarrow ir}$ and $\bm F_{ir \leftrightarrow vis}$ are fed into infrared and visible HIASSI, respectively. Take infrared HIASSI as example, to guarantee that the complementary information is transferred to harmonious one after MRL, we first shuffle all the channels of $\bm F_{vis \leftrightarrow ir}$, and then divide the first half of $\bm F_{vis \leftrightarrow ir}$ by channel as $\bm F_{grp1}$, the rest half as ${\bm F_{grp2}}$. The images $\hat{\bm I}_{grp2}$ and $\hat{\bm I}_{grp2}$ are respectively reconstructed with $\bm F_{grp1}$ and $\bm F_{grp2}$ through decoders $\bm D_1^{rec}$ and $\bm D_2^{rec}$ (composed of $3\times 3$ convolution and $Tanh$ activation function), whose network structures are the same but with different parameters. To encourage $\bm F_{grp1}$ and $\bm F_{grp2}$ to cover more modality-harmonious information, the reconstruction loss is defined as:
\begin{equation}
	\begin{aligned}
		{\ell_{rec1}} &= ||\hat{\bm I}_{grp1} - \bm I_{ir}|{|_1},\\
		{\ell_{rec2}} &= ||\hat{\bm I}_{grp2} - \bm I_{vis}|{|_1},\\
		{\ell_{rec}} &= {\ell _{rec1}} + {\ell _{rec2}}.
	\end{aligned}
\end{equation}

\subsubsection{Structure information preservation}\label{section:3.3.2}
As aforementioned, edge detail information plays an important role in the infrared and visible image fusion since it can facilitate subsequent processing and help in decision making. Therefore, we design SIP module to preserve rich edge features, whose structure is displayed in Fig. \ref{label2}. Firstly, cross attention is performed on both $\bm F_{grp1}$ and $\bm F_{grp2}$ together to obtain interactive and enhanced features. Mathematically, the interactive enhancement process is denoted as:
\begin{equation}
	\begin{aligned}
		{\widetilde{\bm F}_{grp2 \leftarrow grp1}} = {\mathop{\rm softmax}\nolimits} (\frac{{{{\bf{Q}}_{grp2}}{{({{\bf{K}}_{grp1}})}^T}}}{{\sqrt d }}){{\bf{V}}_{grp1}},\\
		{\widetilde{\bm F}_{grp1 \leftarrow grp2}} = {\mathop{\rm softmax}\nolimits} (\frac{{{{\bf{Q}}_{grp1}}{{({{\bf{K}}_{grp2}})}^T}}}{{\sqrt d }}){{\bf{V}}_{grp2}},
	\end{aligned}
\end{equation}
where $\leftarrow$ denotes feature interaction, ${{\bf{Q}}_{grp1}}$, ${{\bf{Q}}_{grp2}} \in {\mathbb{R}^{C \times HW}}$, ${{\bf{K}}_{grp1}}$, ${{\bf{K}}_{grp2}} \in {\mathbb{R}^{C \times HW}}$, and ${{\bf{V}}_{grp1}}$, ${{\bf{V}}_{grp2}} \in {\mathbb{R}^{C \times HW}}$ are the results of $\bm F_{grp1}$ and $\bm F_{grp2}$ passed through feature vectorization operation and $1\times 1$ convolution, respectively. ${\sqrt d }$ is a normalization factor, and $T$ is transpose operation. To further enhance the features, the following operations are performed:
\begin{equation}
	\begin{aligned}
		{\bm F_{grp2 \leftarrow grp1}^{en}} = Conv_{1\times 1}(\widetilde{\bm F}_{grp2 \leftarrow grp1}) \oplus {\bm F_{grp1}},\\
		{\bm F_{grp1 \leftarrow grp2}^{en}} = Conv_{1\times 1}(\widetilde{\bm F}_{grp1 \leftarrow grp2}) \oplus {\bm F_{grp2}}.
	\end{aligned}
\end{equation}

Secondly, we apply two encoders $\bm E_1^{ed}$ and $\bm E_2^{ed}$ composed of $3\times 3$ convolution and $Tanh$ activation function to ${\bm F_{grp2 \leftarrow grp1}^{en}} $ and  ${\bm F_{grp1 \leftarrow grp2}^{en}} $ so as to obtain edge features  $\bm F_{grp2 \leftarrow grp1}^{ed} $ and  $\bm F_{grp1 \leftarrow grp2}^{ed} $, which are expected to capture more edge-related information:
\begin{equation}
	\begin{aligned}
		\bm F_{grp2 \leftarrow grp1}^{ed} = \bm E_1^{ed}(\bm F_{grp2 \leftarrow grp1}^{en}),\\
		\bm F_{grp1 \leftarrow grp2}^{ed} = \bm E_2^{ed}(\bm F_{grp1 \leftarrow grp2}^{en}).
	\end{aligned}
\end{equation}

To ensure that the network can extract rich edge details, $\bm F_{grp2 \leftarrow grp1}^{ed}$ and $\bm F_{grp1 \leftarrow grp2}^{ed}$ are fed into decoders $\bm D_1^{ed}$ and $\bm D_2^{ed}$ (composed of $3\times 3$ convolution and $Tanh$ activation function) to attain the reconstructed images $\hat{\bm I}_{1}^{ed}$ and $\hat{\bm I}_{2}^{ed}$. The gradient loss is performed to force $\bm E_1^{ed}$ and $\bm E_2^{ed}$ to extract rich edge features:
\begin{equation}
	\begin{aligned}
		{\ell_{grad1}}  &=  ||\nabla \hat{\bm I}_1^{ed} - \nabla \bm I_{ir}|{|_1},\\
		{\ell_{grad2}}  &=  ||\nabla \hat{\bm I}_2^{ed} - \nabla \bm I_{vis}|{|_1},\\
		{\ell_{grad}} &= {\ell _{grad1}} + {\ell _{grad2}},
	\end{aligned}
\end{equation}
where $\nabla $ denotes Laplacian gradient operator.

At this point, we can obtain the edge features extracted by SIP, which can be expressed as:
\begin{equation}
	\begin{aligned}
		\bm F^{ed} = \left[{{\bm F_{grp1 \leftarrow grp2}^{ed}},{\bm F_{grp2 \leftarrow grp1}^{ed}}} \right],
	\end{aligned}
\end{equation}
where $\left[ \cdot \right] $ denotes channel-wise concatenation. For infrared and visible SIPHIA, $\bm F_{ir}^{ed}$ and $\bm F_{vis}^{ed}$ denote the edge features from them, respectively. 

Furthermore, since information loss is inevitable in feature extraction, we also add those enhanced features with rich information to the main fusion procedure to ensure more informative fused results. The enhanced feature in SIPHIA can be expressed as: 
\begin{equation}
	\begin{aligned}
		\bm F^{en} = \left[{{\bm F_{grp1 \leftarrow grp2}^{en}},{\bm F_{grp2 \leftarrow grp1}^{en}}} \right].
	\end{aligned}
\end{equation}
$\bm F_{ir}^{en}$ and $\bm F_{vis}^{en}$ denote the enhanced features from infrared and visible SIPHIA, respectively. 

\subsection{Fusion result reconstruction}\label{section:3.4}
To transfer complementary information to harmonious information, we present the above MIT module to achieve mutual feature representation betweeen two complementary modalities images, and two SIPHIAs to ease MIT of transferring information from different modalities, together for more visually pleasant fusion results. As for the fusion process, we concatenate the mutual represented features of different modalities $\bm F_{vis \leftrightarrow ir}$ and $\bm F_{ir \leftrightarrow vis}$ in MIT and those acquired useful harmonious information from infrared and visible SIPHIA, $i.e.$, $\bm F_{ir}^{ed}$, $\bm F_{ir}^{en}$, $\bm F_{vis}^{ed}$, and $\bm F_{vis}^{ed}$, formulated as follows: 
\begin{equation}
	\begin{aligned}
		\bm I_{fused} = \bm D_{fuse}(\left[\bm F_{vis \leftrightarrow ir},\bm F_{ir}^{ed},\bm F_{ir}^{en},\bm F_{ir \leftrightarrow vis},\bm F_{vis}^{ed},\bm F_{vis}^{en} \right]),
	\end{aligned}
\end{equation}
where $\bm D_{fuse}$ stands for the decoder made up of $3 \times 3$ convolution and $Tanh$ activation.

To retain clear edge details and rich texture information in the fused image, we adopt joint gradient loss $\ell _{JGrad}$. The gradient of the fused image is forced to approach the maximum value between the infrared and visible image gradients. $\ell _{JGrad}$ is formulated as: 
\begin{equation}
	\begin{aligned}
		{\ell _{JGrad}} = ||{\mathop{\rm O}\nolimits} (\max (\left| {\nabla {\bm I_{ir}}} \right|,\left| {\nabla {\bm I_{vis}}} \right|)) - \nabla \bm I_{fused}|{|_1},\label{eq12}
	\end{aligned}
\end{equation}
where $\nabla$ is Laplacian gradient operator. $\max(\cdot)$ denotes taking the maximum value. ${\mathop{\rm O}\nolimits} (\left| x \right|) = x$ denotes finding the original gradient value before taking its absolute value.

In addition, to retain the saliency targets from the two images, we introduce the intensity loss, which can be expressed as:
\begin{equation}
	\begin{aligned}
		{\bm \omega_{ir}} &= {\bm S_{{\bm I_{ir}}}} /({\bm S_{\bm I_{ir}}} - {\bm S_{{\bm I_{vis}}}}), {\bm \omega _{vis}} = 1 - {\bm \omega _{ir}}, \hfill \\
		{\ell _{int}} &= ||({\bm \omega_{ir}} \odot {\bm I_{ir}} + {\bm \omega _{vis}} \odot {\bm I_{vis}})-{\bm I_{fused}}|{|_1}, \hfill \label{eq13}\\ 
	\end{aligned}
\end{equation}
where $\bm S_{\bm I_{ir}}$ and $\bm S_{\bm I_{vis}}$ denote saliency matrices of $\bm I_{ir}$ and $\bm I_{vis}$, which can be computed according to references \cite{39,40,41,42}. ${\bm \omega _{ir}}$ and ${\bm \omega _{vis}}$ denote the weight maps for $\bm I_{ir}$ and $\bm I_{vis}$, respectively. $\odot$ denotes element-wise multiplying operation.

The overall reconstruction loss in MIT is computed by
\begin{equation}
	\begin{aligned}
		{\ell _{mit}} = {\ell _{int}} + {\lambda _{JG}}{\ell _{JGrad}}\label{eq14}
	\end{aligned},
\end{equation}
where ${\lambda _{JG}}$ is the hyper-parameter.

For infrared SIPHIA, $\bm F_{ir}^{en}$ is fed into the decoder $\bm D_{ir}^{en}$ (composed of $3\times 3$ convolution and $Tanh$ activation function) to obtain reconstructed image $\hat{\bm I}_{ir}^{siphia}$. We use the content loss similar with Eq. (\ref{eq13}) to ensure abundant valuable information in enhanced features:
\begin{equation}
	\begin{aligned}
		\ell _{content}^{en} &= ||({\bm \omega_{ir}} \odot {\bm I_{ir}} + {\bm \omega_{vis}} \odot {\bm I_{vis}})-{\hat{\bm I}_{ir}^{siphia}}|{|_1}. \hfill \\
	\end{aligned}
\end{equation}
To encourage the restoration of texture details, we model the
gradient distribution and develop an edge loss similar with Eq. (\ref{eq12}) as
\begin{equation}
	\begin{aligned}
		\ell _{edge}^{en} &= ||{\mathop{\rm O}\nolimits} (\max (\left| {\nabla {\bm I_{ir}}} \right|,\left| {\nabla {\bm I_{vis}}} \right|)) - \nabla \hat{\bm I}_{ir}^{siphia}|{|_1}.\label{eq16}\\
	\end{aligned}
\end{equation}

The overall enhancement loss is
\begin{equation}
	\begin{aligned}
		{\ell _{en}} &= \ell _{content}^{en} + {\lambda _{edge}}\ell _{edge}^{en},\label{eq17}
	\end{aligned}
\end{equation}
where ${\lambda _{edge}} $ is the hyper-parameter to balance the content and edge losses. To sum up, the total loss in each SIPHIA can be computed by
\begin{equation}
	\begin{aligned}
		{\ell_{siphia}} = {\ell_{rec}} + {\ell _{grad}} + {\ell _{en}}\label{eq18}.
	\end{aligned}
\end{equation}

\subsection{Mutual promotion training paradigm}\label{section:3.5}
Since the mutual represented features from MIT, and edge details, enhanced features from two SIPHIAs all have great influence on the final fusion results, making them collaborate with each other to achieve mutual reinforcement becomes a vital step. Simply, we divide the whole training process to two circulative phases, namely phaseM and phaseS. 

In the first phaseM, we train MIT to have the basic complementary to harmonious information transfer and fusion ability. In the first phaseS, two SIPHIAs are trained to create impressive supplementary information, $i.e.$ ${\bm F_m^{ed}}$ and ${\bm F_m^{en}}\left( {m \in \left\{ {ir,vis} \right\}} \right)$, which is also the important guarantee of MIT. It is noteworthy that in the next phaseM, MIT realizes fusion based on the transferred features and injected features from two SIPHIAs in the previous phaseS. In this phaseM, we fix the parameters in both two SIPHIAs which have been optimized before and train MIT. As for phaseS training, once the network in MIT get trained, we fix the optimized parameters of MIT trained in earlier phaseM, and take the fused results $\bm I_{fused}$ as “labels”, pulling close “labels” and two enhanced harmonious fusion results $\hat{\bm I}_{ir}^{siphia}$ and $\hat{\bm I}_{vis}^{siphia}$ through Kullback-Leibler Divergence, formulated as:
\begin{equation}
	\begin{aligned}
		{\ell _{inter}} = \operatorname{KL} ({\bm I_{fused}}||\hat{\bm I}_{ir}^{siphia}) + \operatorname{KL} ({\bm I_{fused}}||\hat{\bm I}_{vis}^{siphia})\label{eq19}.
	\end{aligned}
\end{equation}
In this way, the KL Divergence forces networks in two SIPHIAs to learn better edge details and enhanced features injected to the fusion process, getting added information with higher quality. In the next training stage of MIT, the fused results will be naturally improved with the help of better additional information consequently. By now, a virtuous mutual promotion training cycle has been formed. We further demonstrate the pseudo code of our proposed training paradigm as follows.

\begin{algorithm}
	\caption{Mutual Promotion Training Paradigm}\label{Alg:1}
	\begin{algorithmic}
		\STATE {\textbf{Input:}} infrared images $\bm I_{ir}^{\prime}$ and infrared images $\bm I_{vis}^{\prime}$.
		\STATE {\textbf{Output}} the fused image $\bm I_{fused}$
		\begin{flushleft}
		~1: Sample a batch of source data.\\
		~2: Initialize the related encoders, $\bf Q$, $\bf K$, $\bf V$ and decoders in MIT and SIPHIAs.\\
		~3: \textbf{for} \emph{iter}=1, $\cdots$, \emph{Iteration}$_{Maximum}$ \textbf{do}\\
		~4: ~~\textbf{if} iter\% 200 = even number (phaseM) \\
		~5:\quad~~Train MIT. \\
		~6:\quad~~Update $\bm E_{ir}$, $\bm E_{vis}$, $\bf Q_{ir}$, $\bf K_{ir}$, $\bf V_{ir}$, $\bf Q_{vis}$, $\bf K_{vis}$, $\bf V_{vis}$ and\\ \qquad~~$\bm D_{fuse}$ by minimizing the loss in Eq. (\ref{eq14}).\\
		~7:	~~\textbf{else} (phaseS) \\
		~8:\quad~~Load the learned $\bm E_{ir}$, $\bm E_{vis}$, $\bf Q_{ir}$, $\bf K_{ir}$, $\bf V_{ir}$, $\bf Q_{vis}$,\\ \qquad~~ $\bf K_{vis}$, $\bf V_{vis}$.\\
		~9:\quad~~Train two SIPHIAs.\\
		~10:\quad~~Update $\bm D_1^{rec}$, $\bm D_2^{rec}$, $\bf Q_{grp1}$, $\bf K_{grp1}$, $\bf V_{grp1}$, $\bf Q_{grp2}$,\\ \qquad~~ $\bf K_{grp2}$, $\bf V_{grp2}$, $\bm E_1^{ed}$, $\bm E_2^{ed}$, $\bm D_1^{ed}$, $\bm D_2^{ed}$ and $\bm D_{ir}^{en}$($\bm D_{vis}^{en}$) \\ \qquad~~ by minimizing the loss in Eq. (\ref{eq18}) and Eq. (\ref{eq19}).\\
		~10: ~~\textbf{end if}\\
		~11:\textbf{end for}\\
		\end{flushleft}
	\end{algorithmic}
\end{algorithm}

	
		

\section{Experiments}
\subsection{Experimental configurations}
\textbf{Training dataset}: We select 148 image pairs from the RoadScene \cite{9} benchmark as the training set in this paper, containing rich scenes, such as roads, vehicles, and pedestrians. Since such data scale is insufficient to train a network with superior performance, to expand this dataset, we randomly crop the training samples into $120 \times 120$ patches, in which way the whole training set is enlarged to 148000 sample pairs.

\textbf{Testing dataset}: Firstly, we conduct qualitative and quantitative experiments on chosen 38 pairs of testing set from the RoadScene. Moreover, as is known to all that the generalization ability of the deep learning-based method is also a vital indicator to evaluate model performance, we carry out the generalization experiments on 40 image pairs of the TNO \cite{26} and the VOT2020-RGBT \cite{27} datasets and 40 image pairs of the OTCBVS \cite{28} dataset including scenes of busy pathway intersections on the Ohio State University campus to verify the effectiveness and generalization ability of our CHITNet. TNO dataset contains multi-spectral nighttime imagery of various military-relevant scenarios in grayscale. VOT2020-RGBT is a RGB-infrared paired dataset involving loads of video sequences for Visual-Object-Tracking Challenge. 

\textbf{Evaluation metrics}: For quantitative evaluation, six statistical metrics are selected to objectively assess the fusion performance, including Correlation Coefficient (CC) \cite{29}, Entropy (EN) \cite{30}, Gradient-based Fusion Performance (${Q^{AB/F}}$) \cite{31}, Chen-Varshney Metric (${Q^{CV}}$) \cite{32}, Sum of Correlations of Differences (SCD) \cite{33}, and Structural Similarity Index Measure (SSIM) \cite{34}. CC evaluates the degree of linear correlation between the fused image and source images. EN measures the amount of information involved in the fused image on the basis of information theory. ${Q^{AB/F}}$ calculates how much edge information is transferred from source images to the fused image. ${Q^{CV}}$ is based on the human visual system model, using the Sobel operator to extract the edge information of the source images and the fusion result to obtain the edge intensity map G. Smaller ${Q^{CV}}$ implies more in line with human visual perception. SCD reflects the correlation level between information transmitted to the fused image and corresponding source images. SSIM models image loss and distortion. In addition, a fusion algorithm with larger CC, EN, ${Q^{AB/F}}$, SCD, and SSIM indicates better fusion performance.

\subsection{Implementation details}
We train our proposed CHITNet on selected 148 image pairs with data augmentation from the RoadScene dataset. Since our method is a circulative two-stage model, we train each phase with batch size set to 8 and patch size set to $120 \times 120$. $Iteratio{n_{Maximum}}$ is set to 1000. Furthermore, four Adam optimizers \cite{35}(${\beta _1}$ = 0.9, and ${\beta _2}$ = 0.999) responsible for two SIPHIAs and two different MIT stages with initial learning rate of 0.001, which decreases to $10^{-4}$ after 100 epochs and then to $10^{-7}$ after 400 epochs, except for the second MIT phase with initial learning rate of $10^{-5}$ and the same decay rate, are used to optimize two cyclical training phases under the guidance of the interaction loss $\ell _{inter}$. In addition, two hyperparameters ${\lambda _{edge}}$ and ${\lambda _{JG}}$ are set to 20. Our framework is implemented in PyTorch. All experiments are conducted on an Nvidia 3090 GPU. 

Please note that we simply convert source images in RoadScene, VOT2020-RGBT and OTCBVS datasets to gray to achieve fusion. 

\subsection{Ablation studies}
In the above description, we mainly devise two important modules named MIT and SIPHIA to achieve complementary to harmonious information transfer and edge detail preservation. In this section, we give more results under different settings on the framework and training strategy on TNO-VOT dataset, to verify the rationality of our design.
\subsubsection{Mutual information transfer analysis}
\begin{figure}[t!]
	\centering
	\includegraphics[width=0.5\textwidth]{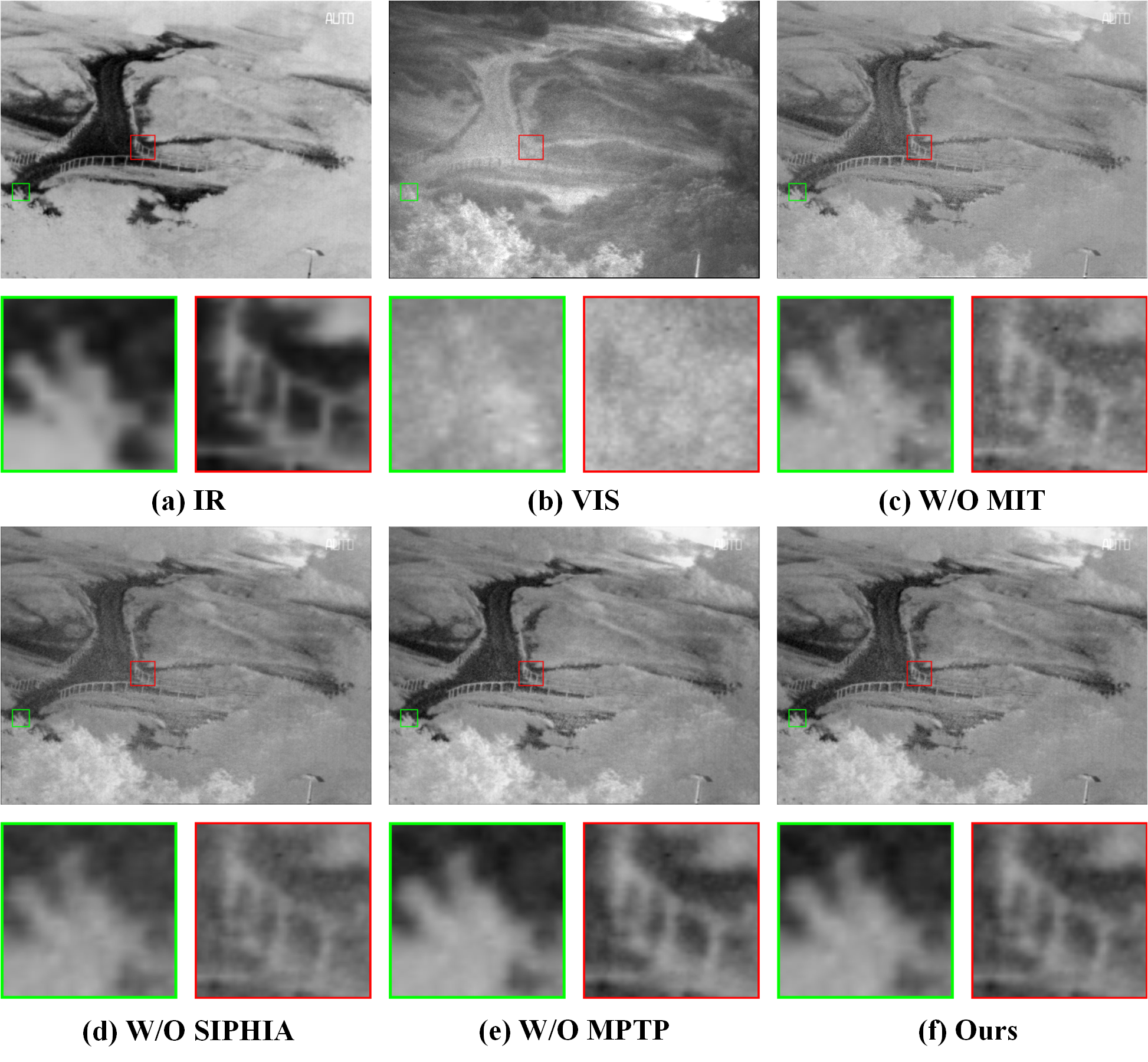}
	\caption{Vision quality comparison of the ablation study on important idea and modules. From left to right, infrared image, visible image, and the results of W/O MIT, SIPHIA, MPTP, and our CHITNet.}
	\label{label3}
\end{figure}

\begin{figure}[t!]
	\centering
	\includegraphics[width=0.5\textwidth]{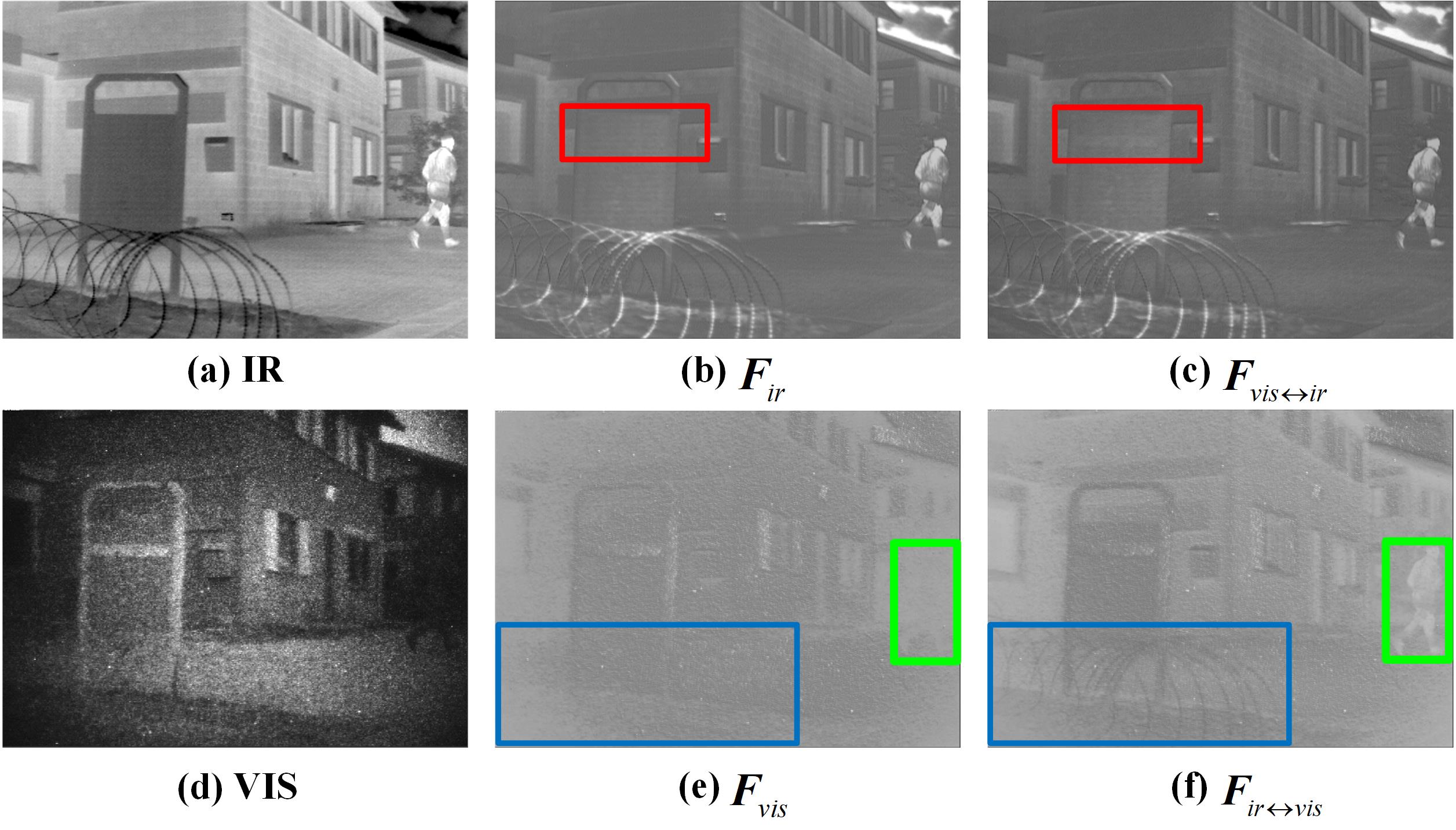}
	\caption{Infrared and visible feature maps before and after mutual information transfer.}
	\label{label4}
\end{figure}

To skillfully achieve IVIF fusion without direct complementary information excavation, we devise the complementary to harmonious information transfer network to concentrate more on harmonious information. Specifically, we design the MIT module to achieve information transfer, as well as two SIPHIAs to help MIT ensure successful transfer process. After we change the whole network into a single feature extraction and fusion one, $i.e.$,  features extracted are modality-specific, the quality of fused image declines slightly. As shown in Fig. \ref{label3}(c), some edge details of the bush and handrail in the infrared image are lost, and the image contrast decreases. Furthermore, so as to demonstrate the effectiveness of the MIT more intuitively, we display the averaged infrared and visible feature maps on channel before and after the MIT module, $i.e.$, $\bm F_{ir}$, $\bm F_{vis \leftrightarrow ir}$, $\bm F_{vis}$ and $\bm F_{ir \leftrightarrow vis}$ in Fig. \ref{label4}. We can clearly see that the strip on the road sign in visible image successfully transfers to the infrared image. Moreover, the wire netting and walking man in the infrared image successfully transfer to the visible image, which implies successful complementary to harmonious information transfer. Besides, the quantitative results in Table \ref{lambda_metrics1} drop the most in practically all evaluation metrics, which speaks volumes for our complementary to harmonious information transfer idea.

\begin{table}[!t]\footnotesize
	\centering {
		\caption{Quantitative evaluation results of ablation study on 40 pairs of infrared and visible images from TNO-VOT dataset.}
		\label{lambda_metrics1}
		\renewcommand\arraystretch{1.2}
		\begin{tabular}{ccccccc}
			\hline
			\hline
			Configuration   &CC &EN &$Q^{AB/F}$ &$Q^{CV}$ &SCD & SSIM \\
			\hline
			w/o MIT     &\it 0.6874  &6.7817   &0.4520 &598.0829 &1.2261 &1.3919 \\
			w/o SIPHIA     &0.6843  &6.9485   &\it 0.4570 &524.1034 &\it 1.3557 &1.3945 \\
			w/o MPTP &0.6813	&\it 7.0756	    &0.4143	&\it 454.8244 &1.3168 &\it 1.4202 \\
			Ours  &\bf 0.6991 &\bf 7.2819 &\bf 0.5503	&\bf 427.2510 &\bf 1.4893 &\bf 1.4443 \\
			\hline
	\end{tabular}}
\end{table}

\begin{table}[!t]\footnotesize
	\centering {
		\caption{Quantitative evaluation results of loss function ablation study on 40 pairs of infrared and visible images from TNO-VOT dataset.}
		\label{lambda_metrics2}
		\renewcommand\arraystretch{1.2}
		{\footnotesize\centerline{\tabcolsep=3.2pt
				\begin{tabular}{ccccccc}
					\hline
					\hline
					Configuration   &CC &EN &$Q^{AB/F}$ &$Q^{CV}$ &SCD & SSIM \\
					\hline
					${\ell_{rec}} $    &\it 0.6913  &6.3758   &0.4088 &462.4592 &1.2261 &1.3828 \\
					${\ell_{rec}}+{\ell_{grad}}$     &0.6491  &\it 6.9219   &\it 0.4580 &\it 456.7497 &\it 1.3670 &\it 1.4219 \\
					${\ell_{grad}}+{\ell_{en}}$     &0.6317  &6.5864   &0.4029 &480.3978 &1.2091 &1.2988 \\
					${\ell_{rec}}+{\ell_{grad}}+{\ell_{en}}$ &\bf 0.6991 &\bf 7.2819 &\bf 0.5503	&\bf 427.2510 &\bf 1.4893 &\bf 1.4443 \\
					\hline
	\end{tabular}}}}
\end{table}

\begin{figure*}[t!]
	\centering
	\includegraphics[width=0.9\textwidth]{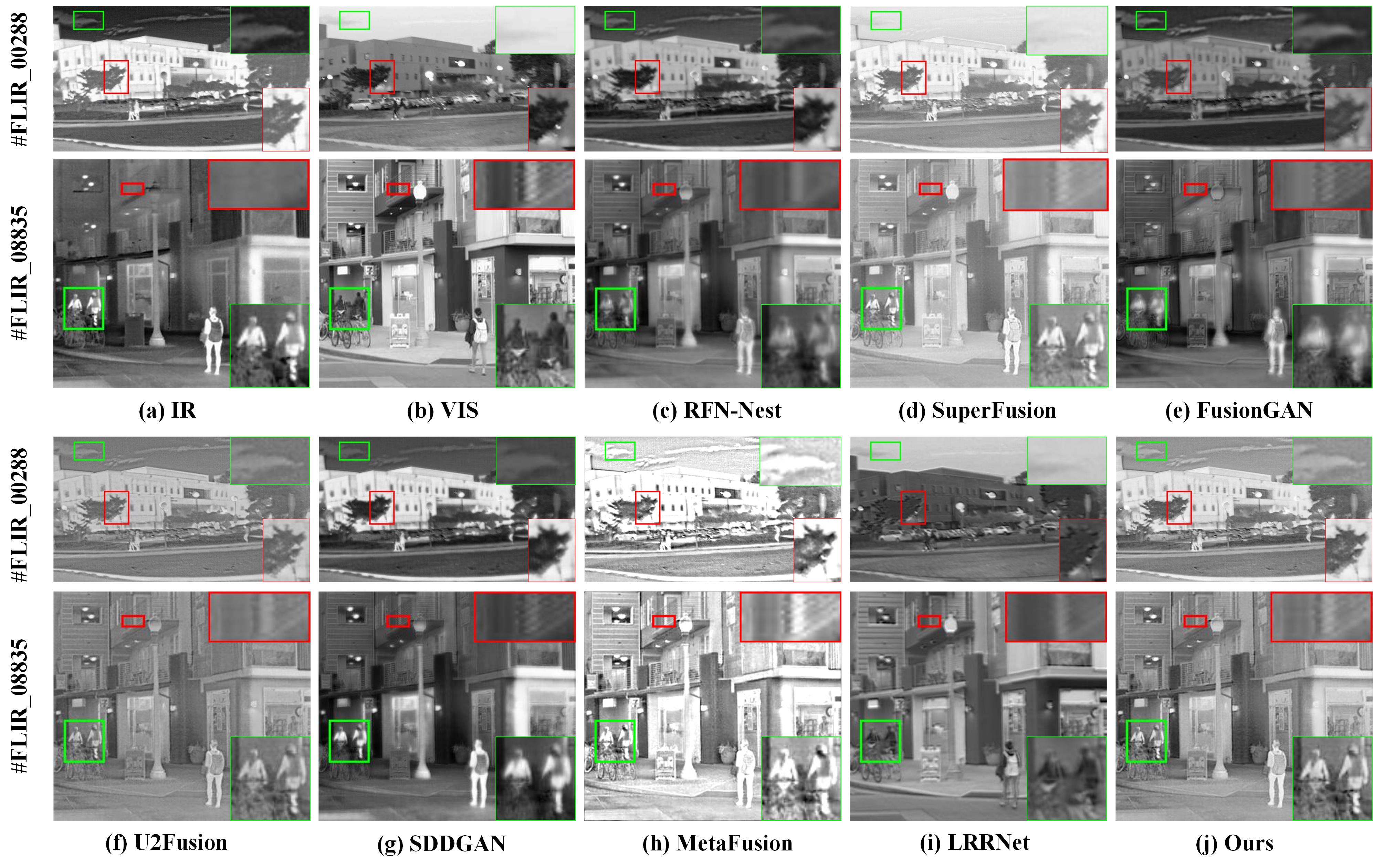}
	\caption{Vision quality comparison of our method with seven SOTA fusion methods on \#FLIR\_00288 and \#FLIR\_08835 images from the RoadScene dataset.}
	\label{label5}
\end{figure*}

\subsubsection{Structure information preserved harmonious information acquisition analysis}
Two SIPHIAs are responsible for the retention and boost of edge and texture, in which HIASSI and SIP modules are also the key points in our overall framework since they can effectively enhance the edge and texture details in the fused images. When we remove both of the two SIPHIAs, the edges of nearly all the objects are indistinguishable, meanwhile the texture details blur, as shown in Fig. \ref{label3}(d). On the contrary, our fusion result presents more detailed textures as shown in Fig. \ref{label3}(f). Since ${Q^{AB/F}}$ measures the amount of edge information that is transferred from source images to the fused image, lower ${Q^{AB/F}}$ in model without SIPHIAs further indicates the effectiveness of the proposed SIPHIA module.
\subsubsection{Mutual promotion training paradigm analysis}
We propose a circulative mutual promotion training strategy to improve the final fusion performance across the board. In the ablation experiment, we adopt end-to-end training paradigm, which is to say, we train two SIPHIAs and MIT together as a whole. The visualized result is presented in Fig. \ref{label3}(e). Though the visualized result is similar to ours, the quantitative results in Table \ref{lambda_metrics1} deteriorate. Therefore, we can draw a conclusion that MPTP provides a more effective training for the fusion process.
\subsubsection{Loss function analysis}
To better analyze the impact of individual and composite objective functions, we carry out ablation study on the adopted loss functions. ${\ell_{rec}}$ is to encourage $F_{grp1}$ and $\bm F_{grp2}$ to cover more modality-harmonious information, ${\ell_{grad}}$ is to force $\bm E_1^{ed}$ and $\bm E_2^{ed}$ to extract rich edge features, ${\ell_{en}}$ is to ensure abundant valuable information and encourage the restoration of texture details in enhanced features. Concretely, we adjust the loss function settings in SIPHIA with only ${\ell_{rec}}$, removing ${\ell_{grad}}$ and ${\ell_{en}}$, only ${\ell_{rec}}$ and ${\ell_{grad}}$, removing ${\ell_{en}}$, only ${\ell_{grad}}$ and ${\ell_{en}}$, removing ${\ell_{rec}}$, and ours. Table \ref{lambda_metrics2} shows the quantitative results, from which we can see that there is a severe decline in CC under configuration without ${\ell_{rec}}$, which indicates that ${\ell_{rec}}$ encourages modality-harmonious information acquisition to a large extent. Furthermore, considerable increase in EN, ${Q^{AB/F}}$, SCD, and SSIM occurs under configuration with ${\ell _{rec}} + {\ell _{grad}}$ compared to the configuration with only ${\ell_{rec}}$, implying that ${\ell_{grad}}$ effectively ensures the structure information preservation. Meanwhile, ${Q^{CV}}$ drops significantly in our model compared with ${\ell _{rec}} + {\ell _{grad}}$ configuration, indicating that ${\ell_{en}}$ helps to enhance texture details and satisfy the human visual perception. The best result comes with the loss function configuration in our model, which means the ${\ell_{rec}}$, ${\ell_{grad}}$ and ${\ell_{en}}$ together promise the effectiveness of the SIPHIA module.
\begin{figure*}[t!]
	\centering
	\includegraphics[width=\textwidth]{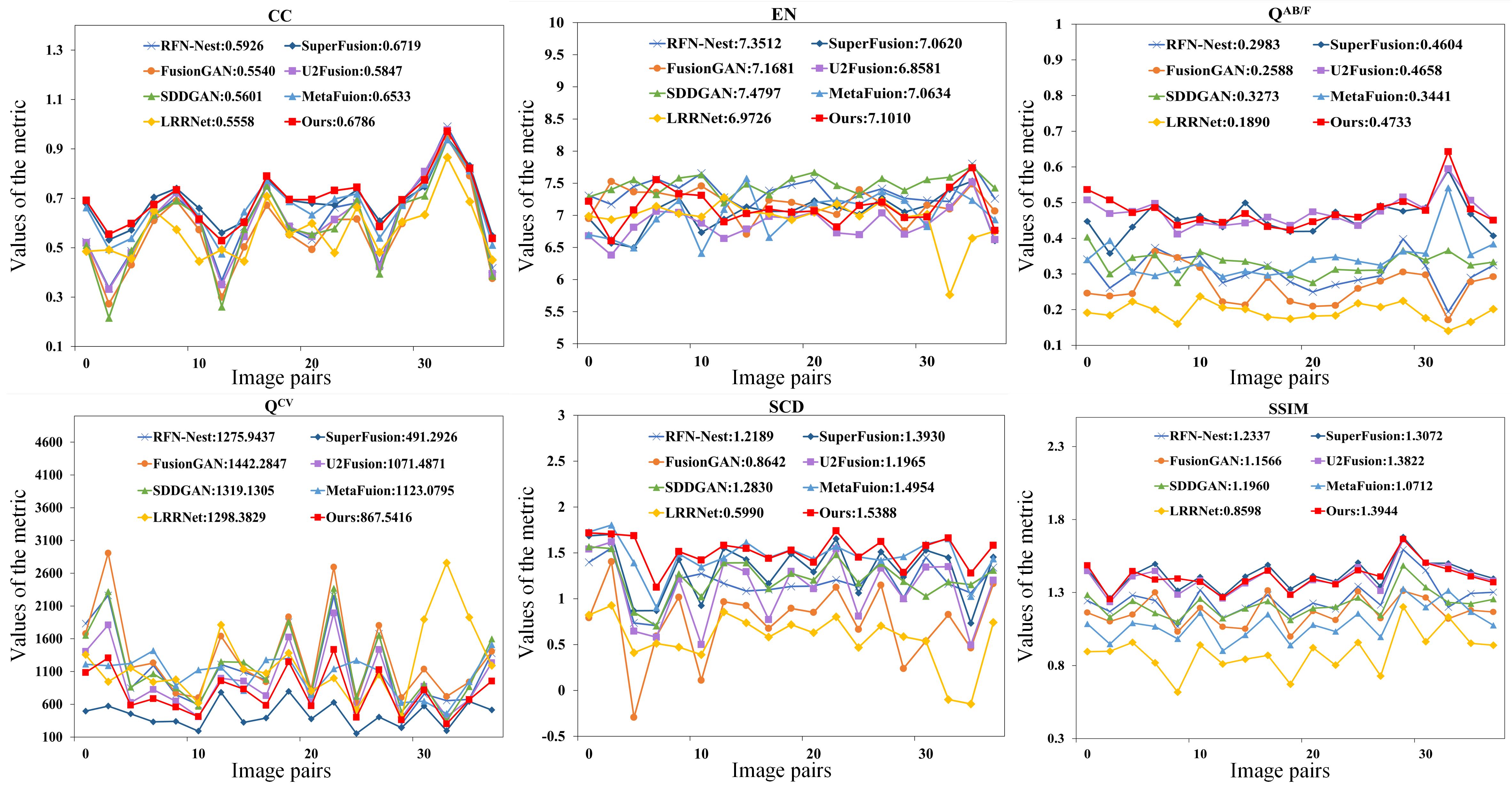}
	\caption{Quantitative results of six metrics, $i.e.$, CC, EN, $Q^{AB/F}$, $Q^{CV}$, SCD, and SSIM, on 38 image pairs from RoadScene dataset. Seven SOTA methods are used for comparison.}
	\label{label6}
\end{figure*}
\begin{table*}[!t]\footnotesize
	\centering {
		\caption{Quantitative results (mean$\pm$standard deviation) of six metrics, $i.e.$, CC, EN, $Q^{AB/F}$, $Q^{CV}$, SCD, and SSIM, on 38 image pairs from RoadScene dataset. Bold: best. Italic: second best.}
		\label{lambda_metrics3}
		\renewcommand\arraystretch{1.2}
		\begin{tabular}{ccccccc}
			\hline
			\hline
			Methods   &CC &EN &$Q^{AB/F}$ &$Q^{CV}$ &SCD & SSIM \\
			\hline
			RFN-Nest\cite{12}  &0.5926$\pm$0.1620   &7.3512$\pm$0.1798   &0.2983$\pm$0.0461 &1275.9437$\pm$566.9294 & 1.2189$\pm$0.2554 & 1.2337$\pm$0.1166 \\
			SuperFusion\cite{51}	&0.6719$\pm$0.1003	&7.0620$\pm$0.2010  	&0.4604$\pm$0.0456	&\bf 491.2926$\pm$478.2370 &1.3930$\pm$0.2786	&1.3072$\pm$0.1143 \\
			FusionGAN\cite{14}   &0.5540$\pm$0.1652 &7.1681$\pm$0.2211	&0.2588$\pm$0.0491 &1442.2847$\pm$640.4603 &0.8642$\pm$0.3980 &1.1566$\pm$0.0954 \\
			U2Fusion\cite{9} 	& 0.5847$\pm$0.1652	&6.8581$\pm$0.2419	&0.4658$\pm$0.0477	& 1071.4871$\pm$569.4802 &1.1965$\pm$0.3573 &\it 1.3822$\pm$0.0986 \\
			SDDGAN\cite{36} &0.5601$\pm$0.1769 &\bf 7.4797$\pm$0.1456
			&0.3273$\pm$0.0485	&1319.1305$\pm$480.8573  &1.2830$\pm$0.2169 &1.1960$\pm$0.0883 \\
			MetaFusion\cite{57}	&0.6533$\pm$0.1144	&7.0634$\pm$0.2321	&0.3441$\pm$0.0441	&1123.0795$\pm$951.0889 &1.4954$\pm$0.2687	&1.0712$\pm$0.0948 \\
			LRRNet\cite{38}	&0.5558$\pm$0.1614	&6.9726$\pm$0.3130  	&0.1890$\pm$0.0337	&1298.3829$\pm$548.8963 &0.5990$\pm$0.2781	&0.8598$\pm$0.1359 \\
			MSID\cite{49}  &0.6283$\pm$0.1422  &7.3218$\pm$0.1997 &0.4679$\pm$0.0417 &834.6983$\pm$451.0366 & 1.4439$\pm$0.2944 & 1.3334$\pm$0.1453 \\
			SwinFusion\cite{52}	&\it 0.6812$\pm$0.1051	&6.6821$\pm$0.4255  	&0.4441$\pm$0.0436	&\it 492.6137$\pm$422.8416 &1.5021$\pm$0.3158	&1.3558$\pm$0.1242 \\
			DATFuse\cite{53}   &\bf 0.6857$\pm$0.1043 &6.6218$\pm$0.4201	&0.4647$\pm$0.0493 &553.8302$\pm$499.5408 &1.2170$\pm$0.2425 &1.3283$\pm$0.1006 \\
			ML-Fusion\cite{56} 	& 0.6725$\pm$0.1109	&7.0620$\pm$0.3152	&0.4349$\pm$0.0458	& 616.9118$\pm$523.6615 &1.2337$\pm$0.2648 &1.3418$\pm$0.1427 \\
			STDFusionNet\cite{8} &0.6753$\pm$0.1299 &6.7231$\pm$0.3669
			&0.4351$\pm$0.0467	&567.1185$\pm$502.1178  &1.3541$\pm$0.3426 &1.3222$\pm$0.1102 \\
			DenseFuse\cite{10}	&0.6116$\pm$0.1482	&7.2331$\pm$0.2874	&0.4623$\pm$0.0533	&1011.2296$\pm$619.9858 &1.1735$\pm$0.2224	&1.2899$\pm$0.1131 \\
			NestFuse\cite{11}	&0.6805$\pm$0.1117	& 7.4260$\pm$0.4120  	&\it 0.4718$\pm$0.0418	&568.6777$\pm$438.0157 &1.2468$\pm$0.2771	&1.2685$\pm$0.1188 \\
			GANMcC\cite{17}	&0.6325$\pm$0.1542	&7.3186$\pm$0.2246  	&0.3180$\pm$0.0422	&1132.8647$\pm$562.5429 &1.0884$\pm$0.3055	&1.2403$\pm$0.1245 \\
			SeAFusion\cite{22}	&0.6783$\pm$0.1312	&\it7.4443$\pm$0.2047  	&0.4608$\pm$0.0474	&496.5111$\pm$422.9348 &\it 1.5266$\pm$0.3019	&1.2376$\pm$0.1160 \\
			Ours  &0.6786$\pm$0.1077 &7.1010$\pm$0.2688 &\bf 0.4733$\pm$0.0398	&867.5416$\pm$330.5593 &\bf 1.5388$\pm$0.2094 &\bf 1.3944$\pm$0.0992 \\
			\hline
	\end{tabular}}
\end{table*}
\subsection{Comparative experiment}
To illustrate the superiority of the proposed method, we conduct comprehensive comparative experiments of fusion performance with sixteen SOTA competitors including RFN-Nest \cite{12}, SuperFusion \cite{51}, FusionGAN \cite{14}, U2Fusion \cite{9}, SDDGAN \cite{36}, MetaFusion\cite{57}, LRRNet \cite{38}, MSID\cite{49}, SwinFusion\cite{52}, DATFuse\cite{53}, MetaLearning-Fusion\cite{56}, STDFusionNet\cite{8}, DenseFuse\cite{10}, NestFuse\cite{11}, GANMcC\cite{17} and SeAFusion\cite{22} on 38 image pairs of RoadScene, 40 image pairs of the TNO and the VOT2020-RGBT, and 40 image pairs of OTCBVS datasets. Due to the limited space and line chart confusion that may be brought by too many compared methods, we only show the qualitative results of seven representative SOTA methods and display the quantitative results of all  sixteen compared SOTA methods in the comparison tables.
\begin{figure*}[t!]
	\centering
	\includegraphics[width=0.9\textwidth]{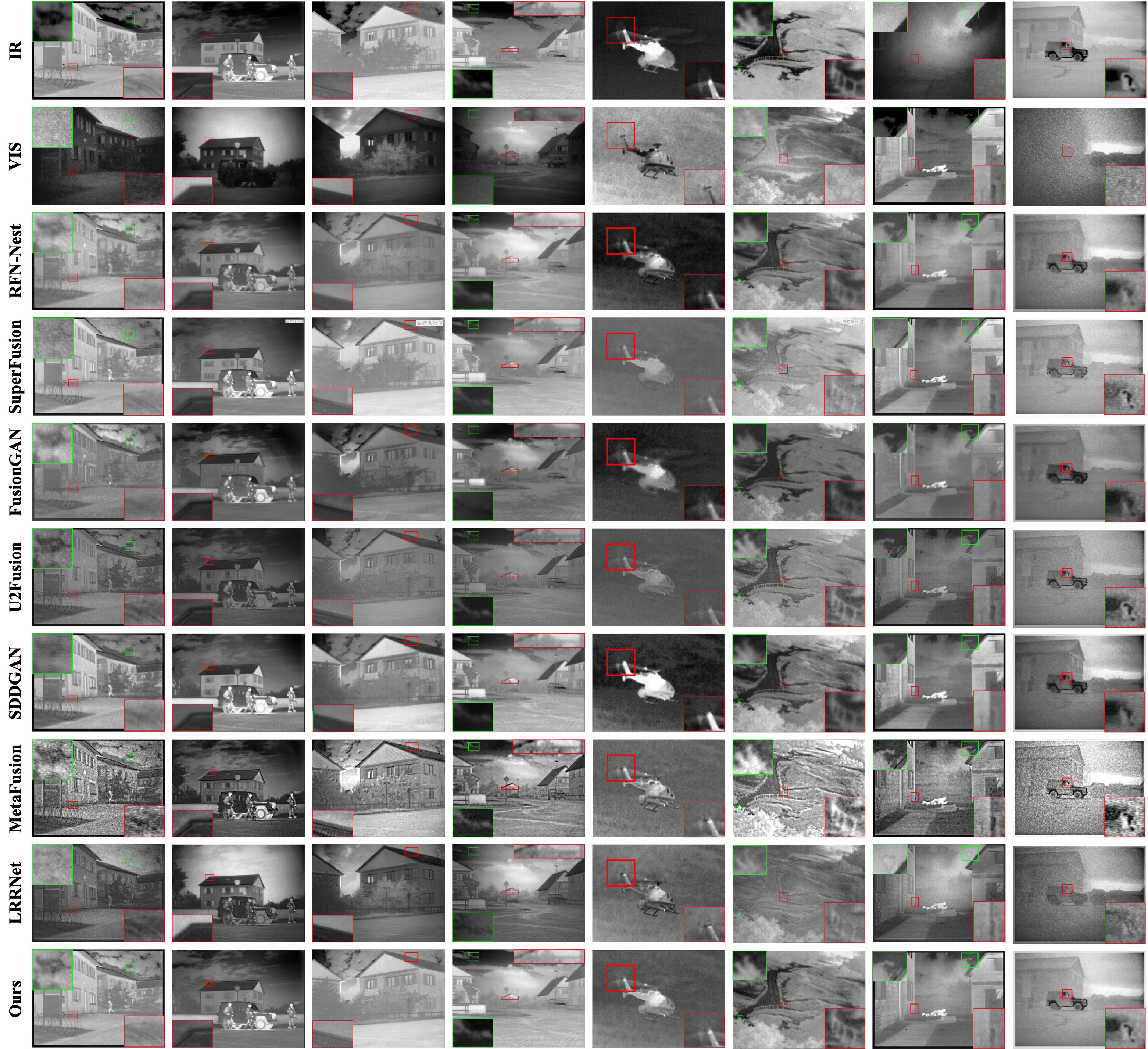}
	\caption{Vision quality comparison of our method with seven SOTA fusion methods on eight images from the TNO-VOT dataset. Each column represents a scene.} 
	\label{label7}
\end{figure*}
\begin{figure*}[t!]
\centering
\includegraphics[width=\textwidth]{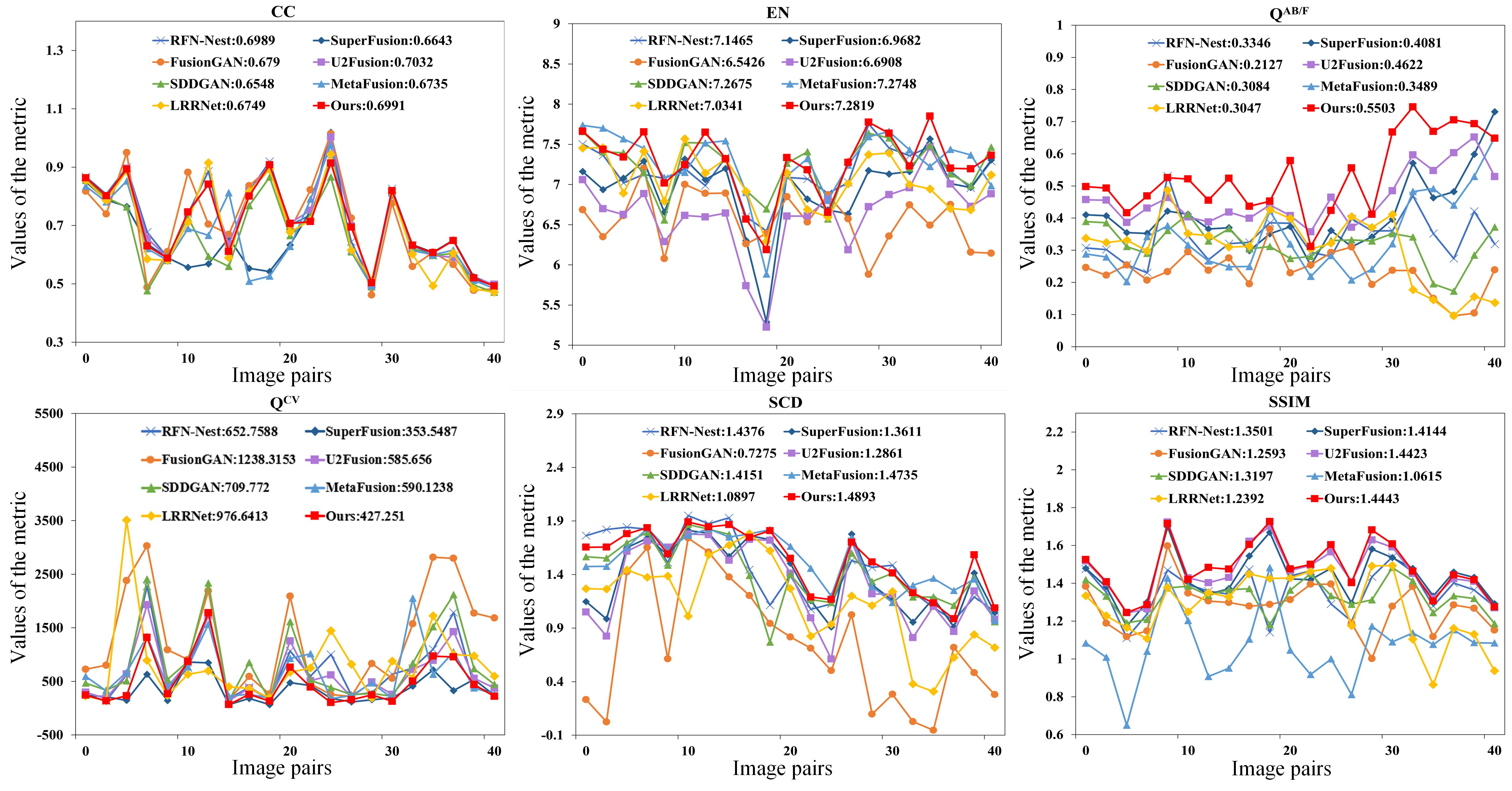}
\caption{Quantitative results of six metrics, $i.e.$, CC, EN, $Q^{AB/F}$, $Q^{CV}$, SCD, and SSIM, on 40 image pairs from TNO-VOT dataset. Seven SOTA methods are used for comparison.}
\label{label8}
\end{figure*}
\begin{table*}[t!]\footnotesize
\centering {
	\caption{Quantitative results (mean$\pm$standard deviation) of six metrics, $i.e.$, CC, EN, $Q^{AB/F}$, $Q^{CV}$, SCD, and SSIM, on 38 image pairs from RoadScene dataset. Bold: best. Italic: second best.}
	\label{lambda_metrics4}
	\renewcommand\arraystretch{1.2}
	\begin{tabular}{ccccccc}
		\hline
		\hline
		Methods   &CC &EN &$Q^{AB/F}$ &$Q^{CV}$ &SCD & SSIM \\
		\hline
		RFN-Nest\cite{12}  &0.6989$\pm$0.1398  &7.1465$\pm$0.3824   &0.3346$\pm$0.0833 &652.7588$\pm$474.2458 &1.4376$\pm$0.2984 & 1.3501$\pm$0.2984 \\
		SuperFusion\cite{51}	&0.6643$\pm$0.1188	&6.9682$\pm$0.4419  	&0.4081$\pm$0.1182	&353.5487$\pm$222.6246 &1.3611$\pm$0.3025	&1.4144$\pm$0.3025 \\
		FusionGAN\cite{14}   &0.6790$\pm$0.1733 &6.5426$\pm$0.3451	&0.2127$\pm$0.0661 &1238.3153$\pm$891.8665 &0.7275$\pm$0.5344 &1.2593$\pm$0.5344 \\
		U2Fusion\cite{9} 	&\bf 0.7032$\pm$0.1426	&6.6908$\pm$0.4399	&0.4622$\pm$0.0854	& 585.6560$\pm$400.8418 &1.2861$\pm$0.3443 &\it 1.4423$\pm$0.3443 \\
		SDDGAN\cite{36} &0.6548$\pm$0.1364 &7.2675$\pm$0.3708
		&0.3084$\pm$0.0704	&709.7720$\pm$555.3076  &1.4151$\pm$0.2651 &1.3197$\pm$0.2651 \\
		MetaFusion\cite{57}	&0.6735$\pm$0.1293	&\it 7.2748$\pm$0.3844	&0.3489$\pm$0.1219	&590.1238$\pm$424.1492 &1.4735$\pm$0.2727	&1.0615$\pm$0.2727 \\
		LRRNet\cite{38}	&0.6749$\pm$0.1474	&7.0341$\pm$0.3848  	&0.3047$\pm$0.1109	&976.6413$\pm$895.7898 &1.0897$\pm$0.4788	&1.2392$\pm$0.4788 \\
		MSID\cite{49}  &0.6632$\pm$0.1345  &7.0562$\pm$0.4052   &0.4702$\pm$0.0988 &\it 348.6385$\pm$300.5214 & 1.3555$\pm$0.3328 &1.3912$\pm$0.3241 \\
		SwinFusion\cite{52}	&0.6271$\pm$0.1368	&6.6821$\pm$0.3955  	&0.3573$\pm$0.1329	&399.4059$\pm$339.5627 &1.3912$\pm$0.4152	&1.2191$\pm$0.3362 \\
		DATFuse\cite{53}   &0.6763$\pm$0.1324 &6.5304$\pm$0.4102	&0.4119$\pm$0.0954 &595.3792$\pm$423.5982 &0.9345$\pm$0.3746 & 1.3796$\pm$0.2985 \\
		ML-Fusion\cite{56} 	& 0.6784$\pm$0.1272	&6.7938$\pm$0.4338	&0.3858$\pm$0.1105	& 452.9399$\pm$412.8850 &1.0498$\pm$0.3950 &1.3473$\pm$0.2993 \\
		STDFusionNet\cite{8} &0.5651$\pm$0.1355 &6.5978$\pm$0.3716
		&0.2258$\pm$0.0977	&665.4629$\pm$450.2568  &0.7460$\pm$0.2996 &1.0667$\pm$0.3620 \\
		DenseFuse\cite{10}	&0.6784$\pm$0.1491	&6.9149$\pm$0.3798	&0.4266$\pm$0.1215	&648.9181$\pm$442.5243 &\it 1.5024$\pm$0.4253	&1.3785$\pm$0.3152 \\
		NestFuse\cite{11}	&0.6267$\pm$0.1370	&7.0914$\pm$0.3223  	&0.5185$\pm$0.1130	&438.0390$\pm$411.5682 &1.1510$\pm$0.3955	&1.3716$\pm$0.3748 \\
		GANMcC\cite{17}	&0.6413$\pm$0.1366	&6.7898$\pm$0.3359  	&0.2615$\pm$0.0901	&962.4854$\pm$544.2882 &0.7834$\pm$0.4156	&1.3040$\pm$0.4156 \\
		SeAFusion\cite{22}	&0.6894$\pm$0.1357	&7.1785$\pm$0.4027  	&\it 0.4713$\pm$0.1143	&\bf 320.0738$\pm$312.0203 &\bf 1.5181$\pm$0.3253	&1.3505$\pm$0.3115 \\
		Ours  &\it 0.6991$\pm$0.1346 &\bf 7.2819$\pm$0.3133 &\bf 0.5503$\pm$0.1050	&427.2510$\pm$351.5149 &1.4893$\pm$0.2734 &\bf 1.4443$\pm$0.2734 \\
		\hline
\end{tabular}}
\end{table*}
\subsubsection{Experiments on RoadScene dataset}
So as to visually evaluate the fusion performance of different algorithms on RoadScene dataset, two pairs of infrared and visible images (named \#FLIR\_00288 and \#FLIR\_08835, respectively) are selected, shown in Fig. \ref{label5}. As illustrated in the green and red boxes in the \#FLIR\_00288 image, our proposed CHITNet keeps the clearest cloud edge while maintaining favourable shrub textures. RFN-Nest, FusionGAN and SDDGAN have vague overall scenes, which declines the visual effects. SuperFusion and LRRNet lose cloud information, and are unable to present satisfying contrast.

As shown in the green box of the \#FLIR\_08835 image, fusion results of RFN-Nest, FusionGAN, SDDGAN, and LRRNet are relatively blurred, missing quite a lot of scene information. Although SuperFusion and U2Fusion have more sharpened pedestrain edges, the railing textures are lost, as zoomed in the red boxes. Oppositely, our fusion results obtain both clear target edges and rich texture details. 

We conduct quantitative comparisons on 38 image pairs from RoadScene dataset to verify the effectiveness of our method, which is presented in Fig. \ref{label6} and Table \ref{lambda_metrics3}. It can be seen that our method ranks first in three metrics. The best SCD and SSIM metrics demonstrate that our results contain more realistic information. The presence of the SIP module allows the ${Q^{AB/F}}$ metric of our method to outperform other SOTA methods. Our method trails SuperFusion by a narrow margin in the ${Q^{CV}}$ metric, but our fusion result preserves more detailed textures and the scene is clearer than that of other methods.

In conclusion, our CHITNet is fully capable of excavating edge and texture features and integrating them into fused images with the help of two SIPHIAs and MIT. Thereby, our method is superior to other SOTA approaches and obtains high-quality fused images.

\subsubsection{Experiments on TNO-VOT dataset}
In order to visually observe the fusion performance of different algorithms on the selected TNO-VOT dataset, we select eight pairs of infrared and visible images. The visualized results are presented in Fig. \ref{label7}. As shown in the first column of Fig. \ref{label7}, nearly all methods have blurred wire netting edges while our method keeps clear edge information, better viewed in the zoomed-in red box. Although SuperFusion performs well in retaining texture details of the wires, it loses clouds textures in the sky, best viewed in the green box. The same situation also happens to the fourth and seventh columns of Fig. \ref{label7}. On the other hand, our method preserves both edge details and background textures to the most extent. 

As depicted in the second and third columns of Fig. \ref{label7}, the highlighted area in the red box indicates that our proposed method could reach the sharpest edges and maintain satisfying brick textures. Moreover, in the sixth column of Fig. \ref{label7}, fusion results of SuperFusion and LRRNet submerge fence into fuzzy background. Although RFN-Nest, FusionGAN, and SDDGAN have relatively clearer edges, they are still indistinct. On the contrary, our method reserves sharpened fence edges as well as considerable texture details.

Finally, as presented in the eighth column of Fig. \ref{label7}, edge details of the saliency target blur in FusionGAN results, largely due to the insufficient capacity in fully modelling the data distributions of source images. Typically, fused results in U2Fusion method lose the thermal target information and cannot attract human attention. Furthermore, existing approaches tend to extract background detail features from the visible image, while acquire salient features from the infrared image. They neglect texture information in the infrared image, resulting partly detail missing. Instead, our proposed method is able to highlight prominent target while preserving favorable contrast and edge details, as well as introducing less noise.

In conclusion, our method could retain the intensity distribution of the infrared target with more sharpened edges and skillfully exclude useless information, $i.e.$ noises in visible images. We attribute this advantage to the fact that our method injects useful edge detail features into the fusion process and spontaneously exploits valuable features in source images.

\begin{figure*}[t!]
	\centering
	\includegraphics[width=0.9\textwidth]{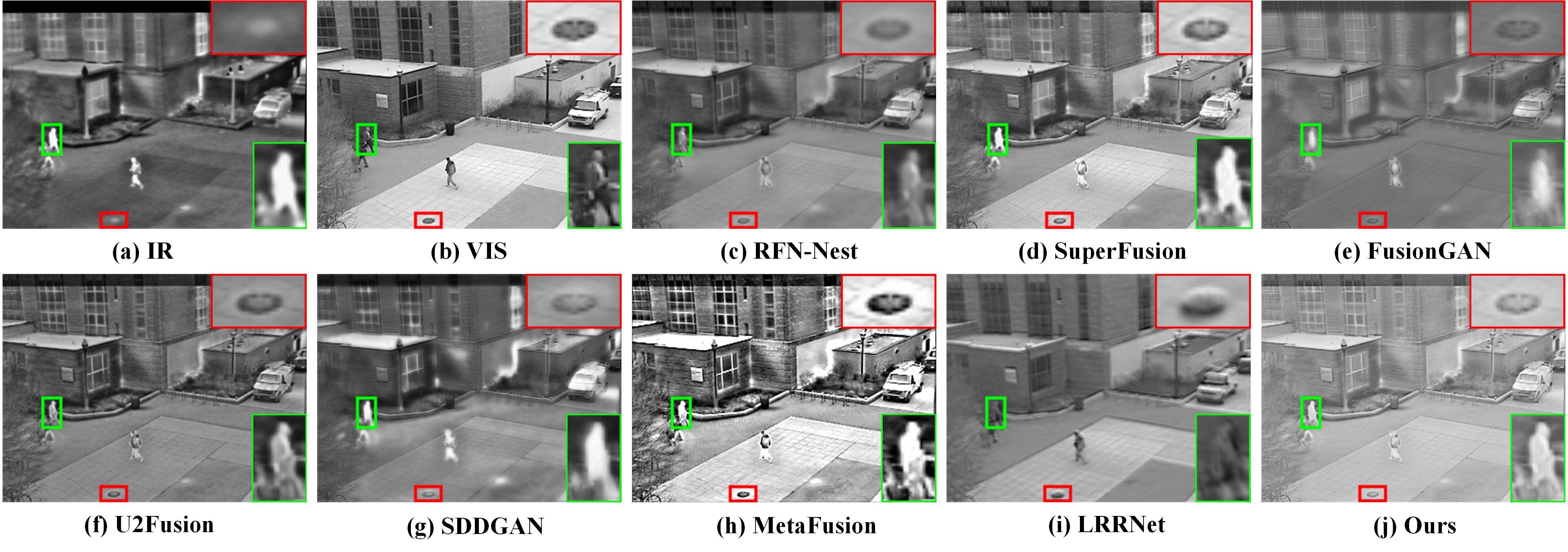}
	\caption{Vision quality comparison of our method with seven SOTA fusion methods on \#video\_0001 from the OTCBVS dataset.}
	\label{label9}
\end{figure*}

\begin{figure*}[t!]
	\centering
	\includegraphics[width=\textwidth]{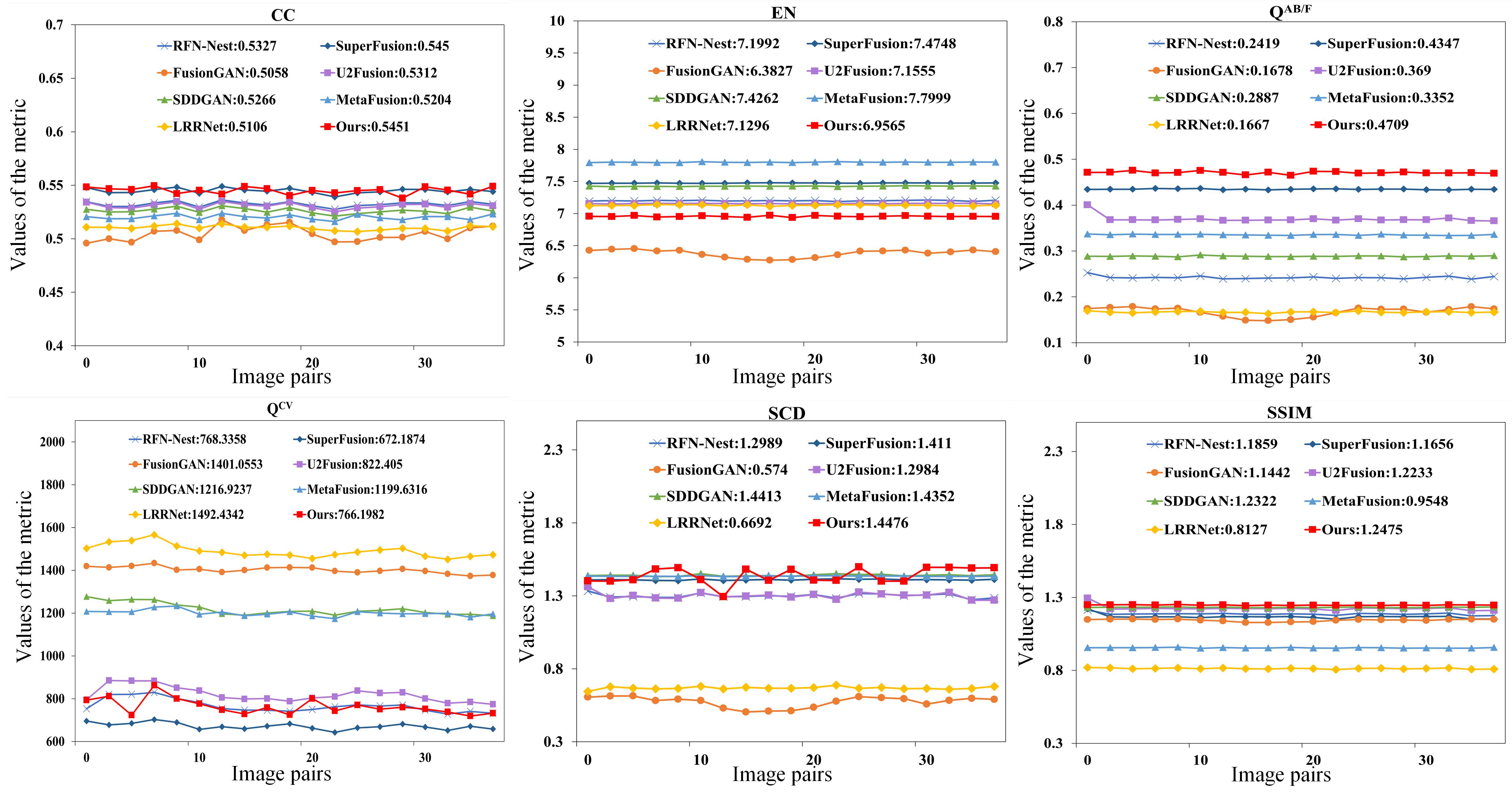}
	\caption{Quantitative results of six metrics, $i.e.$, CC, EN, $Q^{AB/F}$, $Q^{CV}$, SCD, and SSIM, on 40 image pairs from OTCBVS dataset. Seven SOTA methods are used for comparison.}
	\label{label10}
\end{figure*}

\begin{table*}[htbp!]\footnotesize
	\centering {
		\caption{Quantitative results (mean$\pm$standard deviation) of six metrics, $i.e.$, CC, EN, $Q^{AB/F}$, $Q^{CV}$, SCD, and SSIM, on 38 image pairs from RoadScene dataset. Bold: best. Italic: second best.}
		\label{lambda_metrics5}
		\renewcommand\arraystretch{1.2}
				\begin{tabular}{ccccccc}
					\hline
					\hline
					Methods   &CC &EN &$Q^{AB/F}$ &$Q^{CV}$ &SCD & SSIM \\
					\hline
					RFN-Nest\cite{12}  &0.5327$\pm$0.0021  &7.1992$\pm$0.0063   &0.2419$\pm$0.0024 &768.3358$\pm$31.8234 & 1.2989$\pm$0.0132 & 1.1859$\pm$0.0066 \\
					SuperFusion\cite{51}	&0.5450$\pm$0.0022	&7.4748$\pm$0.0065  	&0.4347$\pm$0.0112	&672.1874$\pm$34.1258 &1.4110$\pm$0.0137	&1.1656$\pm$0.0112 \\
					FusionGAN\cite{14}   &0.5058$\pm$0.0067 &6.3827$\pm$0.0595	&0.1678$\pm$0.0102 &1401.0553$\pm$13.5863 &0.5740$\pm$0.0346 &1.1442$\pm$0.0073 \\
					U2Fusion\cite{9} 	& 0.5312$\pm$0.0024	&7.1555$\pm$0.0031	&0.3690$\pm$0.0054	& 822.4050$\pm$37.2177 &1.2984$\pm$0.0191 &1.2233$\pm$0.0133 \\
					SDDGAN\cite{36} &0.5266$\pm$0.0025 &7.4262$\pm$0.0039
					&0.2887$\pm$0.0010	&1216.9237$\pm$25.8822  &1.4413$\pm$0.0049 &1.2322$\pm$0.0018 \\
					MetaFusion\cite{57}	&0.5204$\pm$0.0021	&\bf 7.7999$\pm$0.0064	&0.3352$\pm$0.0056	&1199.6316$\pm$34.1208 &1.4352$\pm$0.0032	&0.9548$\pm$0.0019 \\
					LRRNet\cite{38}	&0.5106$\pm$0.0020	&7.1296$\pm$0.0088  	&0.1667$\pm$0.0012	&1492.4342$\pm$31.3821 &0.6692$\pm$0.0084	&0.8127$\pm$0.0037 \\
					MSID\cite{49}  &0.5345$\pm$0.0033  &\it 7.5968$\pm$0.0238   &\it0.4684$\pm$0.0069 &646.9192$\pm$54.2846 &\it 1.4447$\pm$0.0098 &1.2471$\pm$0.0058 \\
					SwinFusion\cite{52}	&0.5508$\pm$0.0039	&7.5030$\pm$0.0152  	&0.4079$\pm$0.0132	&\it 443.3309$\pm$24.7882 &1.4345$\pm$0.0154	&\bf 1.2561$\pm$0.0026 \\
					DATFuse\cite{53}   &\bf 0.5581$\pm$0.0044 &7.1305$\pm$0.0098	&0.4652$\pm$0.0147 &636.9802$\pm$30.1480 &1.1996$\pm$0.0132 &1.1907$\pm$0.0125 \\
					ML-Fusion\cite{56} 	& 0.5467$\pm$0.0029	&7.4141$\pm$0.0102	&0.3736$\pm$0.0235	& 656.2872$\pm$29.4131 &1.3543$\pm$0.0097 &1.1858$\pm$0.0036 \\
					STDFusionNet\cite{8} &\it 0.5529$\pm$0.0027 &7.1869$\pm$0.0095
					&0.4228$\pm$0.0074	&469.5880$\pm$31.0253  &1.1295$\pm$0.0115 &1.2006$\pm$0.0058 \\
					DenseFuse\cite{10}	&0.5528$\pm$0.0036	&7.1265$\pm$0.0077	&0.3584$\pm$0.0082	&709.4820$\pm$20.1471 &1.2675$\pm$0.0251	&1.2241$\pm$0.0067 \\
					NestFuse\cite{11}	&0.5471$\pm$0.0025	&7.2714$\pm$0.0328 	&0.4650$\pm$0.0099	&527.6546$\pm$15.0256 &1.3631$\pm$0.0189	&1.2153$\pm$0.0057 \\
					GANMcC\cite{17}	&0.5500$\pm$0.0043	&6.6052$\pm$0.0270  	&0.1709$\pm$0.0120	&1047.7786$\pm$25.1454 &0.9918$\pm$0.0088	&1.1962$\pm$0.0156 \\
					SeAFusion\cite{22}	&0.5468$\pm$0.0038	&7.5360$\pm$0.0029  	&0.4648$\pm$0.0057	&\bf 392.9960$\pm$46.0784 &1.4405$\pm$0.0101	&1.2232$\pm$0.0044 \\
					Ours  &0.5451$\pm$0.0037 &6.9565$\pm$0.0092 &\bf 0.4709$\pm$0.0027	&766.1982$\pm$13.9593 &\bf 1.4476$\pm$0.0081 &\it 1.2475$\pm$0.0024 \\
					\hline
	\end{tabular}}
\end{table*}

\begin{figure}[t!]
	\centering
	\includegraphics[width=0.45\textwidth]{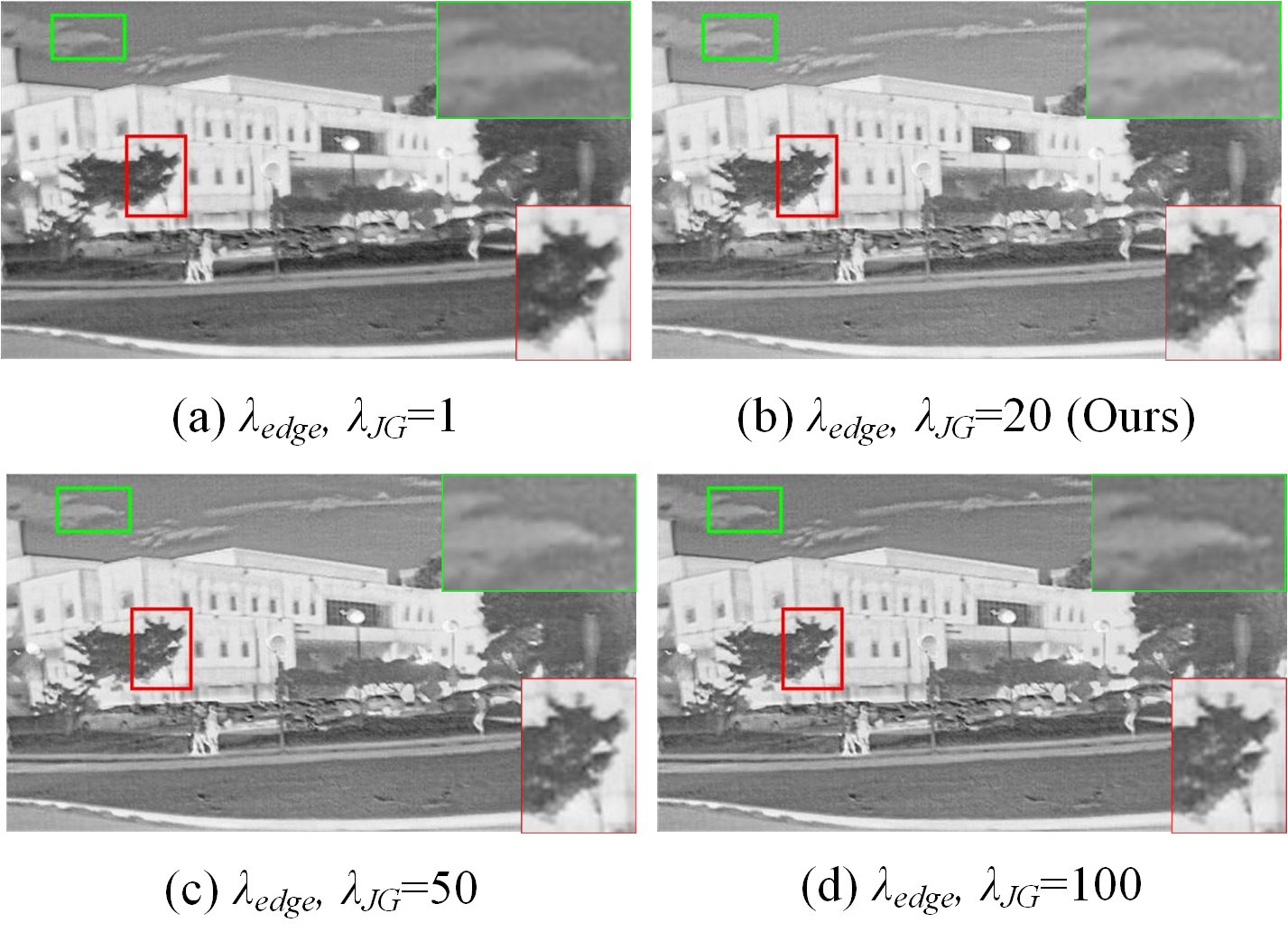}
	\caption{The fusion results with different ${\lambda _{edge}}$ and ${\lambda _{JG}}$.}
	\label{label11}
\end{figure}
The comparative results are presented in Table \ref{lambda_metrics4}. We can see that our method ranks first in three metrics and second in the CC metric. The best EN and SSIM metrics indicate that our results are the most similar with the source images, preserving the original information to the utmost extent and more natural. The idea of complementary to harmonious information transfer promises the advantage of our approach over the SOTA methods. Moreover, our method also performs best in ${Q^{AB/F}}$, showing that our fusion results have richer texture details and more salient targets, with the help of SIP module. Besides, CHITNet has an excellent performance on the ${Q^{CV}}$ metric merely following a few methods, implying that our fused images are also consistent with the human visual system. More intuitive results comparison is shown in Fig. \ref{label8}.

\subsubsection{Experiments on OTCBVS dataset}
Qualitative comparison results on OTCBVS dataset are shown in Fig. \ref{label9} to testify the generalization ability of our CHITNet. Our method is able to provide more sharpened target edges and detailed textures. As shown in the green box of scene \#video\_0001, edge and texture of the walking man with suit is much clearer in our method while other competitors blur to some extent. What's more, though U2Fusion performs neck and neck with ours, it fails to retain the exact shape of the manhole cover, best viewed in the red zoomed-in boxes.

Table \ref{lambda_metrics5} gives quantitatively comparable results, and our CHITNet has three indexes ranking first. The best ${Q^{AB/F}}$ and SCD metrics imply that our results reserve the most information of the source images. More intuitively comparable results are shown in Fig. \ref{label10}.

To sum up, the proposed CHITNet is adept at transferring complementary information to harmonious one, excavating useful information in source images and integrating it into fused images with the help of edge details and enhanced features injection from two SIPHIAs in a co-reinforcement manner. Hence, our method presents more advantages than other SOTA methods and attains high-quality fusion results.

\begin{table}[!t]\footnotesize
	\centering {
		\caption{Effect of different hyperparameters settings on fusion performance.}
		\label{lambda_metrics6}
		\renewcommand\arraystretch{1.2}
		\begin{tabular}{ccccccc}
			\hline
			\hline
			${\lambda _{edge}}$, ${\lambda _{JG}}$   &CC &EN &$Q^{AB/F}$ &$Q^{CV}$ &SCD & SSIM \\
			\hline
			1  &0.6335  &\bf 7.3042   &0.4685 &878.0353 & 1.2989 &\bf 1.4926 \\
			20	 &\bf 0.6786 &7.1010 &\bf0.4733	&\bf 867.5416 &\bf 1.5388 & 1.3944 \\
			50   &0.6181 &7.0921	&0.4361 &932.8368 &1.4078 &1.2952 \\
			100 	& 0.6006 &7.0501 &0.4452 & 987.6895 &1.2984 &1.3956 \\
			\hline
	\end{tabular}}
\end{table}

\subsection{Analysis of hyperparameters}
We perform hyperparameter analysis on RoadScene dataset \cite{9} to validate the rationality of our CHITNet. As described in SIP module of Section \ref{section:3.3.2} and the fusion result reconstruction process in Section \ref{section:3.4}, ${\lambda _{edge}}$ and ${\lambda _{JG}}$ are the parameters controlling the degree of edge loss in the image reconstruction procedure. According to Eq. (\ref{eq17}) and Eq. (\ref{eq14}), the larger ${\lambda _{edge}}$ and ${\lambda _{JG}}$ values pay more attention to the edge maintenance. Since the loss functions for both fused image reconstruction of $\hat{\bm I}_{ir}^{siphia}$($\hat{\bm I}_{vis}^{siphia}$) and $\bm I_{fused}$ are the same, we change those two hyperparameters simultaneously. Now that the edge and texture details directly affect the qualities of fusion results, we assign a large value to ${\lambda _{edge}}$ and ${\lambda _{JG}}$. In order to analyze the sensitivity of this parameter, ${\lambda _{edge}}$ and ${\lambda _{JG}}$ are set as 1, 20, 50, and 100, respectively. Vision and numerical quality results are shown in Fig. \ref{label11} and Table \ref{lambda_metrics6}. It can be seen that when ${\lambda _{edge}}$ and ${\lambda _{JG}}$ are set to 20, the fusion result demonstrates great visual perception and the most metrics reach optimal values.

\begin{table}[!t]\footnotesize
	\centering {
		\caption{Computational efficiency comparison of seven SOTA methods, the value is tested on GPU.}
		\label{lambda_metrics7}
		\renewcommand\arraystretch{1.2}
		{\footnotesize\centerline{\tabcolsep=4.2pt
				\begin{tabular}{ccccccc}
					\hline
					\hline
					Methods   &FLOPs(G) &Size(M) &Time(s) \\
					\hline
					RFN-Nest\cite{12} &- &30.10  &0.36 \\
					SuperFusion\cite{51} &\it 65.43 &0.14 &\it 0.33 \\
					FusionGAN\cite{14}   &497.76 &0.93 	&1.18 \\
					U2Fusion\cite{9} 	&366.34 & 0.66	&1.83 \\
					SDDGAN\cite{36} &561.29 &\bf 0.01 &0.39 \\
					MetaFusion\cite{57}	&- &0.81	&\bf 0.31 \\
					LRRNet\cite{38}	&113.56 &\it 0.05 &0.79	\\
					Ours  &\bf 59.01 &0.88 &0.42 \\
					\hline
	\end{tabular}}}}
\end{table}

\subsection{Analysis of computational complexity}
As shown in Table \ref{lambda_metrics7}, a complexity evaluation is introduced to evaluate the efficiency of our method from three aspects, $i.e.$, FLOPs, training parameters and runtime. Wherein, for FLOPs calculation, the size of the input images is standardized to $512\times 512$ pixels. The inference time is calculated as the average time taken to process 38 scene images from RoadScene’s test dataset. From Table \ref{lambda_metrics7}, our model performs the best in FLOPs, implying that our CHITNet has fast calculation speed and is application-friendly. The average inference time for our model to fuse two source images is 0.42 seconds, only a bit longer than the SOTA method, demonstrating that our model's inference speed is relatively fast and acceptable. Besides, the parameter size of our model is only 0.88M, which can be easily deployed in practical applications. This indicates the efficiency of our CHITNet, which can serve practical vision tasks well with better visual performance.
	
\subsection{Analysis of generalization ability}
\begin{figure}[t!]
		\centering
		\includegraphics[width=0.5\textwidth]{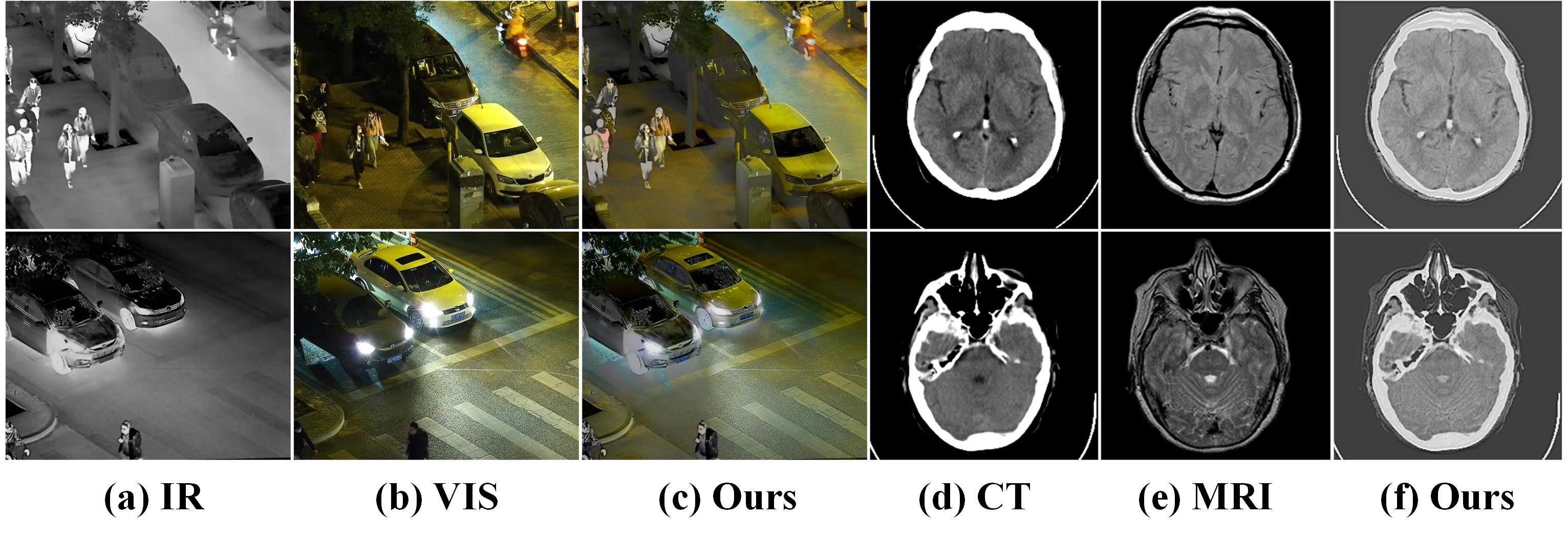}
		\caption{Vision effect of our method on the LLVIP and CT-MRI datasets. (a)--(c) are our fusion results on the LLVIP dataset, and (d)--(f) are our fusion results on the CT-MRI dataset.}
		\label{label12}
\end{figure}

To validate the generalization ability of our method, we conduct experiments on datasets for other image fusion tasks, including LLVIP \cite{62} for color image fusion and CT-MRI \footnote{http://www.med.harvard.edu/AANLIB} for medical image fusion. Fusion results are shown in Fig ~\ref{label12}. From the qualitative results we can see that though our proposed model is committed to IVIF task, it still perfectly completes other fusion tasks.

\subsection{Analysis of limitation}
The proposed method breaks through the limitation of complementary information extraction and aggregation in IVIF task by transferring complementary information into harmonious one, achieving better experimental results compared to other methods. However, our approach still has certain limitations. Specifically, the data required for image processing often undergoes various forms of degradation, such as noise effects. These factors have a negative impact on the robustness of our model. Currently, we have not fully resolved the challenge concerning the model’s limitations to effectively handle these variabilities. Improving the robustness of our method is vital for future research.

\section{Conclusion}
In this paper, we rethink the IVIF task from a brand new angle, focusing more on the easily acquired harmonious information rather than struggling and striving on the complementary information extraction. We propose a novel mutual promoted infrared and visible image fusion algorithm with mutual information transfer and structure information preserved harmonious information acquisition, which achieves effective image fusion with more texture retention. Specifically, MIT fulfills the core idea of complementary to harmonious information transfer, HIASSI is designed to ensure that the complementary information is successfully transferred to harmonious one while SIP is for structure information preservation and enhanced feature injection to avoid vital weak information loss of source images. Then, we employ interaction fusion with the help of added edge details and enhanced features to attain high-quality fused results. Given that components in previous end-to-end training strategies fail to progressively learn from one another, we adopt a mutual promotion training paradigm for better cooperation between each modules. Qualitative and quantitative comparisons with the state-of-the-arts approaches validate the superiority of our method, including edge details, texture information, and visual perception. 

\bibliography{mybibfile}
\end{document}